\pdfoutput=1

\documentclass[11pt]{article}

\usepackage{EMNLP2022}

\usepackage{times}
\usepackage{latexsym}
\usepackage{listings}
\usepackage{xcolor}
\usepackage{color}
\usepackage{latexsym}
\usepackage{graphicx}
\usepackage{booktabs}
\usepackage{algorithm}
\usepackage{algorithmic}
\usepackage{caption}
\usepackage{subfigure}
\usepackage{multirow}
\usepackage{microtype}
\usepackage{amssymb}

\usepackage{footmisc}

\usepackage[T1]{fontenc}

\usepackage[utf8]{inputenc}

\usepackage{microtype}

\usepackage{inconsolata}

\usepackage{microtype}
\usepackage{amsmath, amssymb}
\usepackage{booktabs}
\usepackage{graphicx}
\usepackage{multirow}

\usepackage{bm}
\usepackage{algorithm}
\usepackage{pifont}
\usepackage{subfigure}
\usepackage{xspace}

\title{Less is More: Rethinking State-of-the-art Continual Relation Extraction Models with a Frustratingly Easy but Effective Approach}

\author{
Peiyi Wang$^1$ \quad Yifan Song$^1$ \quad Tianyu Liu$^3$ \quad Rundong Gao$^2$ \\ \textbf{Binghuai Lin$^3$ \quad Yunbo Cao$^3$ \quad Zhifang Sui$^1$}
\\ 
$^1$ Key Laboratory of Computational Linguistics, Peking University, MOE, China \\
$^2$ Center for Data Science, Peking University \\
$^3$ Tencent Cloud Xiaowei \\
 {wangpeiyi9979@gmail.com;}  gaord20@stu.pku.edu.cn \\
 {\{rogertyliu, binghuailin, yunbocao\}@tencent.com;} {\{yfsong, szf\}@pku.edu.cn}
}

\newcommand{\modelname}{\textsc{Fea}\xspace}
\newcommand{\fewrel}{\textsc{Fewrel}\xspace}
\newcommand{\tacred}{\textsc{Tacred}\xspace}
\definecolor{mygreen}{rgb}{0.00,0.70,0.5}
\makeatletter

\newcommand{\Rmnum}[1]{\expandafter\@slowromancap\romannumeral #1@}
\makeatother

\begin{document}
\maketitle
\begin{abstract}
Continual relation extraction (CRE) requires the model to continually learn new relations from class-incremental data streams.
In this paper, we propose a Frustratingly easy but Effective Approach (\modelname) method with two learning stages for CRE: \textbf{1)} Fast Adaption (FA) warms up the model with only new data.
\textbf{2)} Balanced Tuning (BT)
finetunes the model on the balanced memory data.
Despite its simplicity, 
\modelname achieves comparable (on \tacred) or superior (on \fewrel) performance compared with the state-of-the-art baselines.
With careful examinations, we find that the data imbalance between new and old relations leads to a skewed decision boundary in the head classifiers over the pretrained encoders, thus hurting the overall performance. 
In \modelname, the FA stage unleashes the potential of memory data for the subsequent finetuning, while the BT stage
helps establish a more balanced decision boundary.
With a unified view, we find that two strong CRE baselines can be subsumed into the proposed training pipeline. 
The success of \modelname also provides actionable insights and suggestions for future model designing in CRE.

\end{abstract}

\section{Introduction}

Relation extraction (RE) aims to identify the relation of two given entities in a sentence, which is one of the cornerstones in knowledge graph construction and completion \cite{riedel-etal-2013-relation}.
Traditional RE models \cite{gormley-etal-2015-improved, xu-etal-2015-classifying, zhang-etal-2018-graph} are trained on a static dataset with a predefined relation set, which is inadequate in the real-world applications where new relations are constantly emerging.

To adapt to the real-world applications, continual relation extraction (CRE) is proposed \cite{wang-etal-2019-sentence, han-etal-2020-continual}.
CRE requires the model to continually learn new relations from class-incremental data streams.
In CRE, it is often infeasible to combine the new data with previous data and then retrain the model due to the privacy policy, store space limit or computation overhead \cite{biesialska-etal-2020-continual, wu2021curriculum}.
Therefore, CRE models usually suffer from \emph{catastrophic forgetting}, i.e., the performance of previously learned relations drops rapidly while learning new relations.

To alleviate this problem, researchers usually retain a few instances as a memory cache for each learned relation, and propose different ways to replay the sotred 
incorporate the memory data while learning new relations. 
For example, \citet{han-etal-2020-continual} propose memory replay, activation and reconsolidation protocols to attain relation prototype representation.
\citet{cui-etal-2021-refining} propose a memory network to refine the  representation of input instances to better utilize memory data.
\citet{wu2021curriculum} propose a curriculum-meta learning method to reduce the replay frequency of memory data to alleviate the over-fitting problem.
\citet{zhao2022consistent} introduce supervised contrastive learning and knowledge distillation to learn better memory representations.

In this paper, we propose a frustratingly easy but effective approach (\modelname) with two learning stages for CRE:
\textbf{1)}  in \emph{Fast Adaption} (FA), we train the model with only new data to rapidly learn new relations.
\textbf{2)}  in \emph{Balanced Tuning} (BT),
we finetune the model with the balanced updated memory by downsampling new data and then adding the downsampled instances to the existing memory.
Despite its simplicity, \modelname achieves comparable (on \tacred) or superior (on \fewrel) performance compared with the state-of-the-art baselines.

To better understand why our \modelname works, 
we conduct a series of analysis and observe that: 
\textbf{1)} The catastrophic forgetting happens because the CRE models often mistakenly predict existing old relation instances as newly emerging relations.
\textbf{2)} The intrinsic data imbalance problem between newly emerging data and existing memory cache leads to a skewed decision boundary of the head classifier, which is one of the main reasons for catastrophic forgetting.
\textbf{3)} The Fast Adaption phase retains the potential of memory data for the subsequent finetuning, while the Balance Tuning phase circumvents the data imbalance problem, and helps build a more balanced decision boundary.
\textbf{4)} Constructing a better classifier with the balanced decision boundary is of great benefit to CRE models.

With a deep dive into the best performing CRE models, we also find that the proposed \modelname can be viewed as a simplified but more effective variant of two strong baselines, namely EMAR \cite{han-etal-2020-continual} and RPCRE \cite{cui-etal-2021-refining}, by removing or replacing certain modules. 
We show that EMAR and RPCRE can be enhanced by incorporating similar designing principles of \modelname, and the success of \modelname also provides actionable insights and suggestions for future model designing in CRE. 
Our contributions are listed as follows:
\begin{itemize}
    \item We propose a frustratingly easy but effective approach (\modelname) with Fast Adaption and Balanced Tuning for CRE.
    \item Extensive experiments show the effectiveness of \modelname on two benchmarks. We also conduct thorough analysis on why catastrophic forgetting happens and why \modelname works.
    \item By aligning the underlying designs of two strong CRE baselines with that of \modelname, we provide actionable insights for developing effective CRE models in future research.
\end{itemize}

\section{Related Work}
\subsection{Relation Extraction}
Traditional Relation Extraction mainly
includes supervised methods 
\cite{ sup-re-zelenko2003kernel,sup-re-zhou-etal-2005-exploring,zeng-etal-2014-relation,gormley-etal-2015-improved, sup-re-xu-etal-2015-classifying,miwa-bansal-2016-end}, and distant supervised methods \cite{mintz-etal-2009-distant,zeng-etal-2015-distant, lin-etal-2016-neural, han-etal-2018-hierarchical,liu2019reet, sub-re-baldini-soares-etal-2019-matching}.
The conventional RE works mainly utilize human-designed lexical and syntactic features, e.g., shortest dependency path, POS tagging to predict the relation of two entities \cite{sup-re-zelenko2003kernel,sup-re-zhou-etal-2005-exploring,mintz-etal-2009-distant}, which needs a lot of human engineering.
Recently, deep learning-based methods are proposed to alleviate the human labor and outperform feature-based methods.
For example, \citet{zeng-etal-2015-distant} encodes sentence through convolutional neural networks.
\citet{han-etal-2018-hierarchical} introduce attention mechanism to aggregate information of sentence with given entities.
\citet{alt2019improving,sub-re-baldini-soares-etal-2019-matching} introduce pretrained language models for relation extraction.

\subsection{Continual Learning}
Continual Learning (CL) aims to train the models to learn from a continuous data stream \cite{biesialska-etal-2020-continual}.
Researchers usually formulate the data stream as a sequence of tasks arriving at different times.
In CL, the models usually suffer from catastrophic forgetting,
and existing CL methods mainly focus on three kinds of methods to alleviate this problem:
\textbf{1)} Regularization-based methods \cite{regularization-1, regularization-2, regularization-3} put regularization constraints on the parameter space while learning subsequent tasks to preserve acquired knowledge.
\textbf{2)} Parameter isolation-based methods \cite{isolation-1, isolation-2, isolation-3, isolation-4}  allocate subsets of the model parameters or extend new parameters for specific tasks.
For example, \citet{isolation-1} prunes parameters by heuristics and \citet{isolation-4} learns gradient masks to instantiate new sub-networks for each task.
\textbf{3)} Rehearsal-based methods \cite{replay-1, replay-2, replay-3} store a few instances in the memory from previous tasks, and replay the memory data when learning new tasks to remind the model of previously learned knowledge.
Specifically, we focus on the rehearsal-based method for CRE in this paper.

\section{Task Formulation}
Following previous works \cite{cui-etal-2021-refining,zhao2022consistent}, 
CRE is formulated  to accomplish a sequence of tasks $(\mathcal{T}_1, \mathcal{T}_2, \dots, \mathcal{T}_K)$, where the $k$-th task $\mathcal{T}_k$ has its own training set $\mathcal{D}_k$, testing set $Q_k$
and relation set $R_k$.
Every instance $(x_i, y_i) \in \mathcal{D}_k \cup Q_k$ corresponds to a specific relation  $y_i \in R_k$.
More specifically, in the $k$-th task, a CRE model is trained on $\mathcal{D}_k$ to learn new relations $R_k$, and it should also be capable of handling all the seen relations $\bar{R}_k=\cup_{i=1}^kR_i$, i.e., the model will be evaluated on all seen testing sets $\bigcup_{j=1}^iQ_j$.
To circumvent the catastrophic forgetting in CRE, following previous works \cite{cui-etal-2021-refining,zhao2022consistent},
we use a memory to store a few instances $m_r=\{(x_1, r), \dots, (x_B, r)\}$ from each previously learned relation $r$, and then replay them in the subsequent training, where $B$ is the constrained memory size.
After the $k$-th task, we have the memory $\mathcal{M}_k = \cup_{r \in \bar{R}_k}\{(x_1, r), \dots, (x_B, r)\}$, where  $\bar{R}_k$ is the set of all already observed relations.

\section{Methodology}

\renewcommand{\algorithmicrequire}{ \textbf{Input:}} 
\renewcommand{\algorithmicensure}{ \textbf{Output:}} 
\begin{algorithm}[t] 
\caption{\modelname for the $k$-th task} 
\label{alg:easy_approach}
\begin{algorithmic}[1]
\REQUIRE \noindent \\  
$f(\Theta, \theta_{k-1})$: Relation extraction model trained on tasks $(\mathcal{T}_1,..., \mathcal{T}_{k-1})$ with parameters $(\Theta, \theta_{k-1})$. \\
$\mathcal{M}_{k-1}$: Memory which stores typical instances of relations in $(\mathcal{T}_1,..., \mathcal{T}_{k-1})$. \\
$\mathcal{D}_k$: Training dataset of the $k$-th task $\mathcal{T}_k$. \\
\ENSURE \noindent   \\
$f(\Theta, \theta_{k})$: Relation extraction model trained on tasks $(\mathcal{T}_1,..., \mathcal{T}_{k})$ with parameters $(\Theta, \theta_{k})$.   \\
$\mathcal{M}_{k}$: Memory which stores typical instances of relations in $(\mathcal{T}_1,..., \mathcal{T}_{k})$. \\
\STATE{$\theta_k$ $\gets$ $\theta_{k-1} \cup \theta_{new}$}
\FOR{$i\gets 1$ to $epoch_1$}
\STATE{Update $(\Theta, \theta_k)$ with $\nabla \mathcal{L}$ on $\mathcal{D}_k$}
\ENDFOR
\STATE {Select memory instances $m_k$ for relations in $\mathcal{T}_k$ from $\mathcal{D}_k$ through the K\mbox{-}Means algorithm.}
\STATE $\mathcal{M}_k$ $\gets$ $\mathcal{M}_{k-1} \cup m_k$
\FOR{$i\gets 1$ to $epoch_2$}
\STATE{Update  $(\Theta, \theta_k)$  with $\nabla \mathcal{L}$ on $\mathcal{M}_k$}
\ENDFOR
\end{algorithmic}
\end{algorithm}

Our proposed \modelname is model-independent, and we use the relation extraction model $f(\Theta, \theta)$ from \citet{cui-etal-2021-refining} as our backbone model.
The extraction model consists of two components, one is an encoder with parameter $\Theta$, and the other is a classifier with parameter $\theta$.

\subsection{Relation Extraction Model}
\paragraph{Encoder} Given an input instances $x$ with two entities $e_1$ and $e_2$, we first insert four special marks $[E_{11}]$/$[E_{12}]$ and $[E_{21}]$/ $[E_{22}]$ to denote the start/end positions of head and tail entities:$$(\dots,[E_{11}],e_1,[E_{12}],\dots,[E_{21}],e_2,[E_{22}],\dots).$$
Then, we use BERT to encode the input $x$, and get the hidden representations of $[E_{11}]$ and $[E_{21}]$, i.e., $\mathbf{h}_{11}$ and $\mathbf{h}_{21}$.
Finally, we achieve the representation $\mathbf{h}$ of $x$ through as follows:
\begin{equation}
\mathbf{h} = {\rm LayerNorm}(\mathbf{W}_{cat}[\mathbf{h}_{11};\mathbf{h}_{12}] + \mathbf{b}_{cat})), 
\label{eq:hidden}
\end{equation}
where $[;]$ is the concatenation operation, $\mathbf{W}_{cat} \in \mathbb{R}^{d \times 2d}$ and $\mathbf{b}_{cat} \in \mathbb{R}^{2d}$ are trainable parameters.

\paragraph{Classifier}
The classifier figures out the relation probability of $x$ according to the Encoder output  $\mathbf{h}$ as follows:
\begin{equation}
P(y|x) = {\rm softmax}(\mathbf{W}\mathbf{h}),
\end{equation}
where $\mathbf{W} \in \mathbb{R}^{2d \times c}$ is trainable parameter, and $c$ is the number of seen relations.

\begin{table*}[t]
    \centering
    \begin{tabular}{lcccccccccc}
     \toprule
     \multicolumn{11}{c}{\textbf{\textsc{Fewrel}}} \\
     \midrule
        \textbf{Models} & \textbf{T1} & \textbf{T2} & \textbf{T3} & \textbf{T4} & \textbf{T5} & \textbf{T6} & \textbf{T7} & \textbf{T8} & \textbf{T9} & \textbf{T10}  \\
    \midrule
       
       EA-EMR  & 89.0 & 69.0 & 59.1 & 54.2 & 47.8 & 46.1 &  43.1 & 40.7 & 38.6 & 35.2 \\
       CML  & 91.2 & 74.8 & 68.2 & 58.2 & 53.7 & 50.4 & 47.8 & 44.4 & 43.1 & 39.7 \\
        RPCRE  & 97.9 & 92.7 & 91.6 & 89.2 & 88.4 & 86.8 & 85.1 & 84.1 & 82.2 & 81.5  \\
         RPCRE$^{\dagger}$  & 97.8 & 94.7 & 92.7 & 90.3 & 89.4 & 88.0 & 87.1 & 85.8 & 84.4 & 82.8 \\
        EMAR$^{\dagger}$  & 98.1 & 94.3  & 92.3 & 90.5 & 89.7 & 88.5 & 87.2 & 86.1 & 84.8 & 83.6 \\
        CRL   & 98.1 & 94.6 & 92.5 & 90.5 & 89.4 & 87.9 & 86.9 & 85.6 & 84.5 & 83.1 \\
        \midrule
        \modelname (Ours) & \bf{98.3} & \bf{94.8} & \bf{93.1} & \bf{91.7} & \bf{90.8} & \bf{89.1} & \bf{87.9} & \bf{86.8} & \bf{85.8} & \bf{84.3} \\
      
     \bottomrule
     \toprule
    \multicolumn{11}{c}{\textbf{\textsc{Tacred}}} \\
    \midrule
         \textbf{Models} & \textbf{T1} & \textbf{T2} & \textbf{T3} & \textbf{T4} & \textbf{T5} & \textbf{T6} & \textbf{T7} & \textbf{T8} & \textbf{T9} & \textbf{T10}  \\
        \midrule
        EA-EMR  & 47.5 & 40.1 & 38.3 & 29.9 & 24.0 & 27.3 &  26.9 & 25.8 & 22.9 & 19.8\\
        CML & 57.2 & 51.4 & 41.3 & 39.3 & 35.9 & 28.9 & 27.3 & 26.9 & 24.8 & 23.4 \\
        RPCRE & 97.6 & 90.6 & 86.1 & 82.4 & 79.8 & 77.2 & 75.1 & 73.7 & 72.4 & 72.4 \\
        RPCRE$^{\dagger}$  & 97.5 & 92.2 & 89.1 & 84.2 & 81.7 & 81.0 & 78.1 & 76.1 & 75.0 & 75.3 \\
        EMAR$^{\dagger}$ & \bf{98.3} & 92.0 & 87.4 & 84.1 & 82.1 & 80.6 & 78.3 & 76.6 & 76.8 & 76.1 \\
        CRL & 97.7 & \bf{93.2} & \bf{89.8} & 84.7 & 84.1 & 81.3 & 80.2 & \bf{79.1} & \bf{79.0} & \bf{78.0} \\
        \midrule
        \modelname (Ours) & 97.6 & 92.6 & 89.5 & \bf{86.4} & \bf{84.8} & \bf{82.8} & \bf{81.0} & 78.5 & 78.5 & 77.7 \\
    \bottomrule
    \end{tabular}
    \caption{Accuracy on all observed relations at the stage of learning current tasks. $^{\dagger}$ denotes our reproduced results with the open codebases. Other results are directly taken from \citet{zhao2022consistent}. Best results are in \textbf{boldface}. We report the average accuracy over $5$ different runs.
    }
    \label{tab:main}
\end{table*}

\subsection{Two Stage Learning}
In CRE, after the $(k$-$1)$-th task of continual relation extraction, we have the model $f(\Theta, \theta_{k-1})$ which has been trained on $(\mathcal{T}_1, \dots, \mathcal{T}_{k-1})$.
When learning new relations on the $k$-th task, we take a frustratingly easy but effective approach (\modelname) with fast adaption and balanced tuning to train the model.

\paragraph{Fast Adaption (FA)} aims to rapidly learn the knowledge of new tasks.
In FA, as shown in Algorithm \ref{alg:easy_approach}, we first extend the classifier for new relations (line 1), and
then train the model with only the instances of new relations (lines 2-4) to quickly learn new relations. 

\paragraph{Memory Selection}  
After the Fast Adaption, for relations in the current task $\mathcal{T}_k$, we select their most informative and diverse instances from the training data $\mathcal{D}_k$ to update the memory (Algorithm \ref{alg:easy_approach}: line 5-6).
Following \citet{han-etal-2020-continual, cui-etal-2021-refining, zhao2022consistent}, we apply the K-means algorithm to cluster instances of each relation $r$ in $\mathcal{T}_k$ and select the instances closest to each cluster center as the memorized instances of the $r$. The cluster number of K-means is set as the memory size for each relation.

\paragraph{Balanced Tuning (BT)} aims to train the model to distinguish old and new relations.
However, in CRE, the number of instances of old relations in the memory is significantly less than that of current relations in $\mathcal{D}_k$, which causes severe the data imbalance problem.
Therefore, at the Balanced Tuning stage (Algorithm \ref{alg:easy_approach}: line 7-9), we train the model only with the updated memory $\mathcal{M}_k$ where all seen relations have an equal number of instances.

\subsection{Training and Inference}
The loss function of both FA and BT stages is defined as follows:
\begin{equation}
    \mathcal{L}=\sum^{|\mathcal{D}^*|}_{i=1}-\log{P(y_i|x_i)},
\end{equation}
where $(x_i,y_i)$ is an instance from $\mathcal{D}^*$, and $\mathcal{D}^*$ denotes $\mathcal{D}_k$ or $\mathcal{M}_k$ at the FA stage or BT stage, respectively.
During inference, we select the relation with the max probability as the predicted relation.

\section{Experiments}
\subsection{Datasets and Evaluation Metric}
\paragraph{Datasets}
Following \citet{han-etal-2020-continual,cui-etal-2021-refining,zhao2022consistent}, we evaluate \modelname on two widely used datasets, \fewrel and \tacred. 
\textbf{\fewrel} \cite{han-etal-2018-fewrel} is originally proposed for few-shot relation extraction, which consists of $100$ relations and $700$ instances per relation. Previous works \cite{han-etal-2020-continual,cui-etal-2021-refining,zhao2022consistent} construct CRE dataset using $80$ relations of \textsc{Fewrel}, and divide them into $10$ tasks to form a task sequence.
The training, validation and test split ratio is 3:1:1.
\textbf{\tacred} \cite{zhang-etal-2017-position} is a relation extraction dataset with $42$ relations (including the `no relation') and $106,264$ instances. 
\citet{cui-etal-2021-refining, zhao2022consistent} remove the instances of `no relation' and divide the left relations into $10$ tasks to form a task sequence.
The number of training and testing samples for each relation is limited to $320$ and $40$, respectively. 

\paragraph{Evaluation Metric} 
Following \citet{han-etal-2020-continual, cui-etal-2021-refining}, after training on each new task, the model will be evaluated on the test data of all seen relations by the classification accuracy.

\subsection{Experimental Setup}
\paragraph{Parameter Settings}
We use a random sampling strategy to construct continual relation task sequences.
It randomly divides all relations of the dataset into $10$ sets to simulate $10$ tasks.
For fairness, we use the same random seeds with \citet{cui-etal-2021-refining, zhao2022consistent}, thus the task sequences are exactly the same.
We report the average accuracy of $5$ different sampling task sequences.
Following \citet{cui-etal-2021-refining}, we use bert-base-uncased as our encoder, and Adam as our optimizer with the learning rate $1e$-$3$ for non-BERT modules and $1e$-5 for the BERT module.
The memory size is $10$ in our main experiments (we also explore the influence of memory size in Appendix \ref{app:mem}).
Both our fast adaption learning and balanced tuning stages contain $10$ epochs.\footnote{We do not tune the hyperparameters of \modelname, and just set them the same as \citet{cui-etal-2021-refining}.}
We run our code on a single NVIDIA A40 GPU with 48GB of memory.

\paragraph{Baselines}
We compare \modelname with the following baselines in our experiments: \textbf{EA-EMR} \cite{wang-etal-2019-sentence}, \textbf{CML} \cite{wu2021curriculum}, \textbf{EMAR} \cite{han-etal-2020-continual}, \textbf{RP-CRE} \cite{cui-etal-2021-refining}, \textbf{CRL} \cite{zhao2022consistent}. Please refer to appendix \ref{app:baseline} for details of these baselines.

\subsection{Main Results}
The performances of \modelname and baselines are shown in Table \ref{tab:main}.
As is shown, 
Our \modelname achieves the best result at all task (T1-T10) stages on \fewrel, and comparable results with the best performing model CRL on \tacred.
We think \modelname can achieve the new state-of-the-art on \fewrel, since \fewrel is a class-balanced dataset, which is harmonious with our training data distribution of BT.
Although \modelname does not achieve the best performance on \tacred, it is just slightly worse than a more complex model CRL $0.3$ accuracy (Note that \modelname outperforms CRL 1.2 accuracy on \fewrel).
These results show the effectiveness of our \modelname on both balanced and imbalanced datasets.\footnote{We also explore the influence of memory size (the number of clusters in memory selection) for \modelname. \modelname is more stable than baselines. Please refer to Appendix \ref{app:mem} for details.}\footnote{\modelname also has a better model efficiency than that of baselines. Please refer to Appendix \ref{app:me} for more details.}

\section{Analysis}

\begin{table}[t]
\centering
\small
\scalebox{1}{
    \begin{tabular}{lcc}
    \toprule
    \textbf{Methods}  & \textbf{\fewrel} & \textbf{\tacred}  \\
    \midrule
    \modelname & 84.3 & 77.7   \\
     \midrule
    \textit{remove} BT  &  75.8 & 71.2 \\
     \textit{remove} FA   & 81.1 & 74.9 \\
    \textit{remove} FA and BT  & 75.6 & 71.2 \\
    \bottomrule
    \end{tabular}
}
\caption{Performances of models with different training data at two learning stages.}
\label{tab:ablation}
\end{table}

\subsection{Ablation Study}
To explore the effectiveness of \modelname, we conduct the ablation study.
We consider the three different ablation methods: \textbf{1)} \textit{remove} BT, which trains the model on previous memory and all new data (i.e., $\mathcal{M}_{k-1} \cup \mathcal{D}_k$) at the second stage.
\textbf{2)} \textit{remove} FA, which trains the model on  $\mathcal{M}_{k-1} \cup \mathcal{D}_k$ at the first stage.
\textbf{3)} \textit{remove} FA and BT, which directly trains the model on  $\mathcal{M}_{k-1} \cup \mathcal{D}_k$ on each task stage.
As shown in Table \ref{tab:ablation}:
\textbf{a)} \modelname significantly outperforms ``\textit{remove} BT'' $8.5$ and $6.5$ accuracy on \fewrel and \tacred, respectively, which shows BT is very essential for \modelname.
\textbf{b)} ``\textit{remove} FA'' from \modelname leads to $3.2$ and $2.8$ accuracy drop on \fewrel and \tacred, respectively, while ``\textit{remove} FA and BT'' has comparable performance with ``\textit{remove} BT'', showing the effectiveness of FA depends on the presence of BT.
In fact, the FA works since it preserves the potential of memory data for the following BT, which we explore in Section \ref{sec:FA}.
We can conclude that both BT and FA are essential for \modelname.\footnote{We also explore more ablation methods in Appendix \ref{app:ablation}.}

\begin{table}[t]
\centering
\small
\scalebox{1}{
    \begin{tabular}{lccccc}
    \toprule
    \textbf{Methods}  & \textbf{Error} & \textbf{latter}  & \textbf{former}  & \textbf{inner} \\
    \midrule
    \modelname &  15.7\% & 63.5\% & 28.6\% & 7.9\%   \\
    \midrule
    \textit{remove} BT & 24.2\% & 90.3\% & 6.1\% & 3.6\%  \\
    \textit{remove} FA & 18.9\% & 80.9\% & 13.4\% & 6.3\%\\ 
    \midrule
    CRL & 16.9\% & 65.2\% & 28.0\% & 6.8\% \\
    EMAR & 16.6\% & 70.0\% & 23.1\% & 6.9\%  \\
    RPCRE & 17.2\% & 65.2\% & 28.2\% & 6.6\% \\
    \bottomrule
    \end{tabular}
}
\caption{Error analysis on \fewrel. \modelname outperforms other models, mainly since it has fewer \textit{latter errors}. }
\label{tab:la2fo-fo2la}
\end{table}

\subsection{Error Analysis}
\label{sec:error}
The main results and ablation study show \modelname with FA and BT is extremely effective. To understand why \modelname works, we first perform an error analysis to explore where \modelname outperforms other methods.
Specifically, given an input instance $x$ that belongs to the relation $y$ and the corresponding model prediction $y^p$,
we say that the model makes an error if $y \ne y^p$.
Assuming that the relation $y$ and $y^p$ appear at the tasks $t$-th and $t^p$-th, respectively, we summarize three different errors according to their appearance order: \textit{latter error} (i.e., $t < t^p$), \textit{former error} (i.e., $t > t^p$) and \textit{inner error} (i.e., $t = t^p$). 

As shown in Table \ref{tab:la2fo-fo2la},
the \textit{latter error} accounts for a large proportion for all methods.
However, \modelname has less \textit{latter error} proportion compared with other methods and thus outperforms them.
In addition, through the error analysis of \modelname and two ablation methods, we find that both FA and BT are very essential to reduce the \textit{latter error}.
We also conduct a more detailed error analysis and notice that most of the \textit{latter error} happens among similar relations.
For example, on \fewrel, ``\textit{remove} FA'' and ``\textit{remove} BT'' predicts $37.3\%$ and $52.1\%$ instances of the former relation ``location'' to the latter similar relation ``headquarters location'', while the ratios are just $21.1\%$ and $7.8\%$ for \modelname and a supervised model (i.e., trains all data together without forgetting), respectively.
From the error analysis, we can find that {\emph{the catastrophic forgetting happens because the CRE models often mistakenly predict existing old relation instances as newly emerging relations.}}.

\subsection{Explorations of Balanced Tuning}
\begin{figure}[t]
    \centering
    \includegraphics[width=1\linewidth]{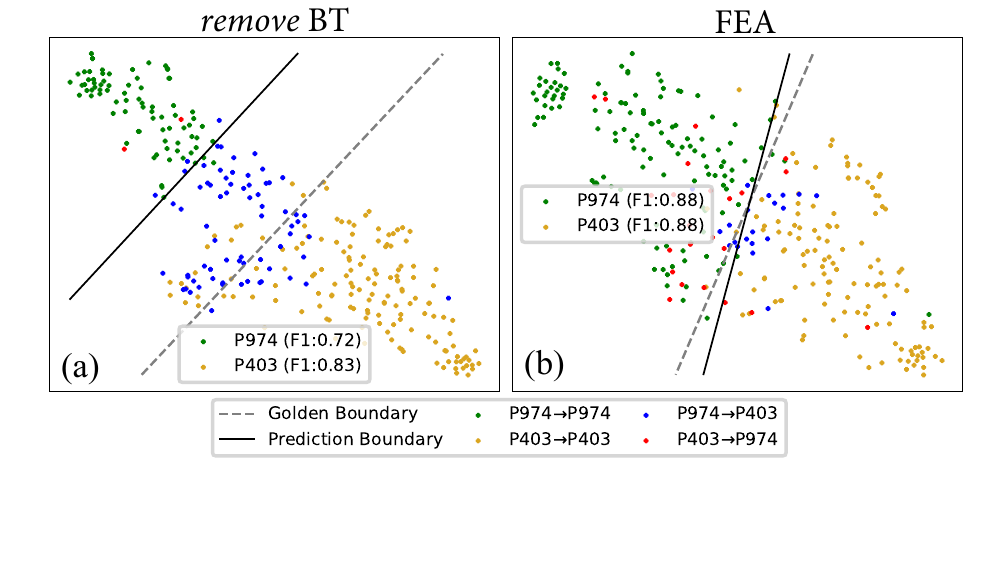}
    \caption{\textit{t-SNE} of instances belonging to P974 (\textit{tributary})  and P403 (\textit{mouth of the watercourse}) after learning P403. ``R1 $\rightarrow$ R2'' means the model predicts the instance of R1 to R2. ``Golden Boundary'' and ``Prediction Boundary'' are the decision boundary computed by the logistic regression with the golden label and model prediction, respectively.}
    \label{fig:BT}
\end{figure}

Compared with ``\textit{remove} BT'', \modelname reduces a lot of \textit{latter error}s, and thus alleviates the catastrophic forgetting.
To explain why BT can reduce the \textit{latter error}, we chose two similar relations P974 (\textit{tributary}, appears at the $7$-th task) and P403 (\textit{mouth of the watercourse}, appears at the $9$-th task) with severe \textit{latter error} to conduct a case study.\footnote{Please refer to Appendix \ref{app:bt} for more cases.}
We draw the \textit{t-SNE} of instances belonging to P974 and P403 for  ``\textit{remove} BT'' and \modelname after learning P403.
As shown in Figure \ref{fig:BT}(a),
for ``\textit{remove} BT'', the model tends to establish a skewed decision boundary between P974 and P403 due to the data imbalance.\footnote{The skewed boundary phenomenon between minor and majority classes is also observed in imbalanced learning works \cite{8953234,9081988}.}
In contrast, \modelname with BT can establish a more balanced decision boundary (Figure\ref{fig:BT}(b)), and thus reduces a lot of \textit{latter error}.
To further confirm whether BT helps build a better decision boundary,
we drop the classifier of \modelname and ``\textit{remove} BT'' after training, and retrain an upper-bound classifier (UBC) with all training data to see their performance gap.
As shown in the first four rows of Table \ref{tab:gold-classifier}:
\textbf{1)} \modelname outperforms ``\textit{remove} BT'', while ``\modelname \textit{with} UBC'' and ``\textit{remove} BT \textit{with} UBC'' have comparable results on two benchmarks, showing that {\emph{BT works because it establishes a more balanced decision boundary}.}
\textbf{2)} \modelname significantly outperforms ``\textit{remove} BT'' average $7.5$ accuracy on two benchmarks, showing that {\emph{the intrinsic data imbalance problem between old and new data that leads to a skewed decision boundary is one of the main reasons for catastrophic forgetting.}}

We also explore the performance of supervised method that trains all data together without catastrophic forgetting. As shown in Table \ref{tab:gold-classifier}:
\textbf{1)} two CRE models ``\textit{with} UBC'' and ``SUP.'' significantly outperform ``SUP. \textit{with} frozen encoder'', which shows the original BERT representation is not good enough to represent relations, and the pretrained encoder learns a lot of knowledge during training.
\textbf{2)} ``SUP.'' significantly outperforms \modelname and ``\textit{remove} BT''  $6.0$ and $13.5$ average accuracy on two benchmarks, while it just outperforms ``\modelname \textit{with} UBC'' and ``\textit{remove} BT \textit{with} UBC'' $2.0$ and $2.1$ average accuracy, showing that  {\emph{the forgetting of BERT encoder may be not a serious problem for the performance of CRE models.}}
\textbf{3)} ``\textit{with} UBC'' significantly improves the performances of \modelname and ``\textit{remove} BT'' $4.0$ and $11.2$ average accuracy on two benchmarks, showing that {\emph{constructing a good classifier can be of great benefit to CRE models.}}

\begin{table}[t]
\centering
\small
\scalebox{1}{
    \begin{tabular}{lccc}
    \toprule
    \textbf{Methods}  & \textbf{\textsc{Few.}} & \textbf{\textsc{Tac.}} & \textbf{Avg.}  \\
    \midrule
    \modelname & 84.3 & 77.7 & 81.0 \\
    \modelname \textit{with} UBC & 88.0 & 82.0 & 85.0 \\
    \midrule
    \textit{remove} BT & 75.8 & 71.2 & 73.5 \\
    \textit{remove} BT \textit{with} UBC & 87.9 & 82.4 & 85.2 \\
    \midrule
    Supervised (SUP.) & 89.5 & 84.5 & 87.0 \\
    SUP. \textit{with} frozen encoder & 77.3 & 66.2 & 71.8 \\
    \bottomrule
    \end{tabular}
}
\caption{Performances of two kinds of supervised methods and \modelname/``\textit{remove} BT'' with upper-bound classifier (UBC) on two benchmarks. A better classifier significantly improves the performance of CRE models.}
\label{tab:gold-classifier}
\end{table}

\subsection{Explorations of Fast Adaption}

\label{sec:FA}

\begin{figure}[t]
    \centering
    \includegraphics[width=1\linewidth]{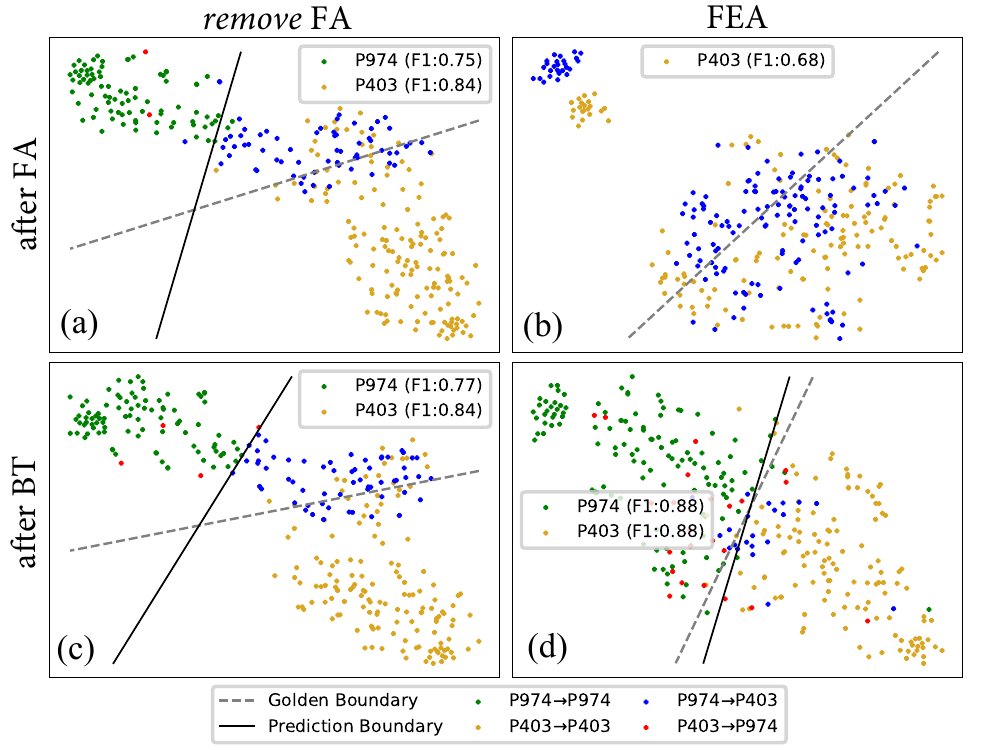}    \caption{\textit{t-SNE} of instances belonging to P974 and P403 after two stages of the task that contains P403.}
    \label{fig:tsne}
\end{figure}

\modelname outperforms ``\textit{remove} FA'' $3.0$ average accuracy on two benchmarks, showing the importance of training the model ONLY on new data at first.
Table \ref{tab:la2fo-fo2la} shows FA works since it is helpful for reducing the \textit{latter error}.
Therefore, we begin with a case study between P974 and P403 with severe \textit{latter error} to understand why FA is helpful.\footnote{Please refer to Appendix \ref{app:FA} for more cases.}
We draw the \textit{t-SNE} of instances belonging to P974 and P403 for  ``\textit{remove} FA'' and \modelname after two learning stages of the task containing P403. 
For ``\textit{remove} FA'',
after the first stage, the model establishes a skewed decision boundary due to the data imbalance problem (Figure \ref{fig:tsne}(a)).
However, after BT, the decision boundary is still skewed  (Figure \ref{fig:tsne}(c)), showing BT does not work well.
For \modelname, in contrast,
after FA, the memory instances of P974 and P403 scatter on the embedding space, and the model predicts all instances as P403 (Figure \ref{fig:tsne}(b)), because the model does not need to distinguish these two relations.
After BT, the model establishes a relative fair decision boundary (Figure \ref{fig:tsne}(d)) compared with that of ``\textit{remove} FA''.
These results show that ``\textit{remove} FA'' may hinder the following BT.

\begin{figure*}[t]
    \centering
    \includegraphics[width=0.9\linewidth]{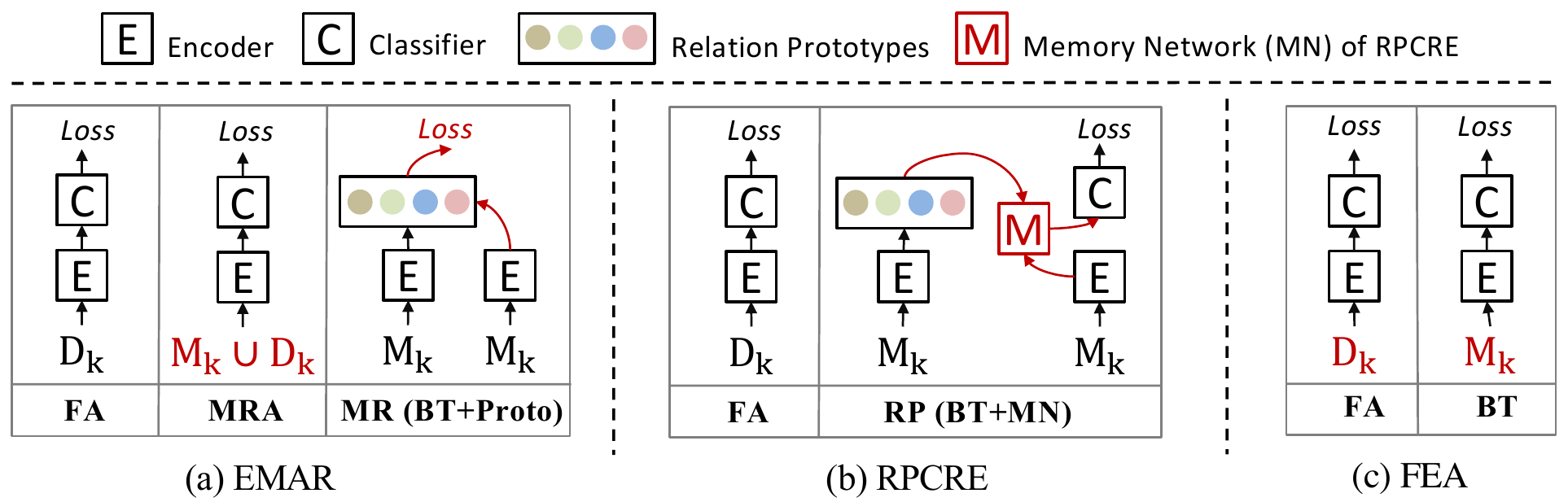}
    \caption{Illustration of EMAR, RPCRE and \modelname. \modelname can be viewed as a simplified variant of EMAR and RPCRE. 
    }
    \label{fig:overview}
\end{figure*}

Because BT updates the model on the memory data, we further utilize the gradient norm of memory data to reflect their effect on BT.
As shown in Table \ref{tab:grad}, in the BT stage,
compared with \modelname, the memory data has a much smaller gradient norm  in ``\textit{remove} FA'', which shows the memory data has a very limited effect.
We think the reasons are as follows: 
\textbf{1)} For ``\textit{remove} FA'', the model has learned a good parameter to  distinguish the memory data between old and new relations at the first stage, and thus tends to keep the learned parameter (i.e., skewed decision boundary) in the following BT.
\textbf{2)} In contrast, \modelname utilizes the memory data only in BT, and thus can help the model establish a more balanced decision boundary on the balanced situation from the scratch.
These results show that {\emph{it is important to learn new relations with only new data at first, and FA works as it retains the potential of memory data for the subsequent BT.}}

\subsection{Less is More: Rethinking CRE models}

In this section, we find that \modelname can be viewed as a simplified variant of two strong CRE models (EMAR and RPCRE), and we rethink them to explain why \modelname can outperform these more complex models.
We hope our analysis can provide guidance for the design of future CRE models.

\paragraph{Rethinking EMAR}
\citet{han-etal-2020-continual} proposes an episodic memory activation and reconsolidation (EMAR) method for CRE.
As shown in Figure \ref{fig:overview}(a), EMAR consists of three stages: \textbf{1)} Learning from new data. \textbf{2)} Memory Replay and Activation (MRA): Learning from the memory data and new data.
\textbf{3)} Memory Reconsolidation (MR): learning from memory data with relation prototypes.
Please refer to \citet{han-etal-2020-continual} for more model details.

Among these three stages, the first stage is the same as FA, and the third stage MR can be regarded as BT (compared with BT, MR further incorporates relation prototypes to train the model).
However, the extra second stage MAR mixes the memory and new data to train the model, and thus introduces the data imbalance problem.
According to our analysis, the data imbalance harms the model performance. Therefore, we only use the balanced memory data to train the model in MRA.
As shown in Table \ref{tab:less-more}, after using balanced data in MRA (EMAR: \textit{balanced} MRA), the accuracy of EMAR on \fewrel and \tacred can improve $0.4$ and $1.1$, respectively.
However, the performance is still slightly lower than that of \modelname, which shows incorporating the relation prototypes may not be helpful in CRE.
As the relation prototypes are calculated by the memory data, we think the reason is that optimizing the model based on relation prototypes aggravates the over-fitting of memory data.

\begin{table}[t]
\centering
\small
\scalebox{1}{
    \begin{tabular}{lcc}
    \toprule
    \textbf{Methods} & \textbf{\fewrel} & \textbf{\tacred} \\
    \midrule
    \modelname  & 27.8 & 45.0  \\
    \textit{remove} FA  & 1.4 &  4.1   \\
    \bottomrule
    \end{tabular}
}
\caption{The average gradient norm per step  of memory data in BT on two benchmarks.} 
\label{tab:grad}
\end{table}
\begin{table}[t]
\centering
\small
\scalebox{1}{
    \begin{tabular}{lcc}
    \toprule
    \textbf{Methods} & \textbf{\fewrel} & \textbf{\tacred}  \\
    \midrule
    \modelname & 84.3   & 77.7   \\
    \midrule
    EMAR  & 83.6   & 76.1   \\
    EMAR: \textit{balanced} MRA  & 84.0  & 77.2  \\
    \midrule
    RPCRE & 82.8  & 75.3 \\
    RPCRE: \textit{with} normalized  & 83.0  & 76.5 \\
 
    \bottomrule
    \end{tabular}
}
\caption{Performances of two typical baseline models after our modification with principles of FEA.} 
\label{tab:less-more}
\end{table}

\paragraph{Rethinking RPCRE}
\citet{cui-etal-2021-refining} proposes a refining network with relation prototypes for CRE (RPCRE).
As shown in Figure \ref{fig:overview}(b), RPCRE consists of two learning stages: \textbf{1)} Initial training for new task. \textbf{2)} Refine input instance with Prototypes (RP): Train a memory network (MN) to refine the representation of input instances with the relation prototypes for better classification.
Please refer to \citet{cui-etal-2021-refining} for more model details.

The first stage is FA, and the second stage RP can be regarded as BT.
Compared with BT, RP further utilizes a memory network (MN) to refine the representation of the input instances with relation prototypes.
Specifically, for each input instance, MN regards it as a query, the relation prototypes as key and value.
MN uses an attention mechanism to aggregate the relation prototypes, and adds the aggregation result to the original representation as a refined representation for classification.

However, MN is harmful for performance.
There exists two potential reasons:
\textbf{1)} MN incorporates implicit data imbalance.
We find that there exists gaps among the norms of different prototypes and input instances tend to attention more to the relation prototype with the larger norm.
Therefore, after refining, the representation of each relation is not balanced.
To alleviate this problem, we normalize the norm of all relation prototypes to $1$ (``\textit{with} normalized'').
As shown in Table \ref{tab:less-more}, after normalizing, the accuracy of RPCRE on \fewrel and \tacred can improve $0.2$ and $1.2$, respectively.
\textbf{2)} MN makes it more difficult for the CRE model to distinguish similar relations.
According to our error analysis in Section \ref{sec:error}, CRE models are confused by the similar relations, and are prone to \textit{latter error}.
Given an instance, the attention-based MN mechanism tends to add the representation of its similar relation prototypes to it, and thus the representation of similar relations becomes more confusing.
For example, RPCRE predicts $40.7\%$ instances of former relation ``followed by'' to latter relation ``follow'', and $5.7\%$ vice versa,
while the ratio of pure \modelname is much less than that of RPCRE ($30.7\%$ and $1.4\%$, respectively).

\subsection{Suggestions for Future Work}
Through our a series of analysis on both \modelname and two strong CRE baselines, we think future CRE models should pay attention to the following issues:
\textbf{1)} \emph{Learn distinguishable feature for similar relations}.
In CRE,  the catastrophic forgetting mainly happens among similar relations from different tasks, since there does not exist enough data to teach the model to distinguish them.
Therefore, it may be helpful to propose some mechanisms to learn distinguishable features that highlight differences among similar relations.
\textbf{2)} \emph{Establish a better classifier}.
In CRE, the forgetting that happens on the pretrained encoder is not serious, and the conventional \textit{Softmax} classifier tends to build a skewed decision boundary, leading to severe \textit{latter errors}.
Therefore, it can be helpful to design a better classifier for CRE models.

\section{Conclusion}
In this paper, we propose a frustratingly easy but effective approach (\modelname) for CRE.
\modelname consists of two stages:  
\textbf{1)} Fast Adaption (FA) warms up the model with only new data.
\textbf{2)} Balanced Tuning (BT) circumvents the intrinsic imbalance between new and old relations by finetuning on the balanced memory data.
Despite the simplicity of \modelname, it is extremely effective.
Therefore, we conduct a series of analysis to understand why \modelname works and why catastrophic forgetting happens.
We also dive into two strong CRE baselines, and find that \modelname can be viewed as a simplified but most effective variant of them.
We show these two CRE baselines can be further enhanced by the principle of \modelname.
Based on our analysis, we also provide two actionable suggestions for future model design in CRE.


\bibliography{emnlp2022}

\begin{thebibliography}{36}
\expandafter\ifx\csname natexlab\endcsname\relax\def\natexlab#1{#1}\fi

\bibitem[{Aljundi et~al.(2018)Aljundi, Babiloni, Elhoseiny, Rohrbach, and
  Tuytelaars}]{regularization-2}
Rahaf Aljundi, Francesca Babiloni, Mohamed Elhoseiny, Marcus Rohrbach, and
  Tinne Tuytelaars. 2018.
\newblock \href
  {https://openaccess.thecvf.com/content_ECCV_2018/html/Rahaf_Aljundi_Memory_Aware_Synapses_ECCV_2018_paper.html}
  {Memory aware synapses: Learning what (not) to forget}.
\newblock In \emph{Proceedings of the European Conference on Computer Vision
  (ECCV)}, pages 139--154.

\bibitem[{Alt et~al.(2019)Alt, H{\"u}bner, and Hennig}]{alt2019improving}
Christoph Alt, Marc H{\"u}bner, and Leonhard Hennig. 2019.
\newblock \href {https://arxiv.org/abs/1906.03088} {Improving relation
  extraction by pre-trained language representations}.
\newblock \emph{arXiv preprint arXiv:1906.03088}.

\bibitem[{Baldini~Soares et~al.(2019)Baldini~Soares, FitzGerald, Ling, and
  Kwiatkowski}]{sub-re-baldini-soares-etal-2019-matching}
Livio Baldini~Soares, Nicholas FitzGerald, Jeffrey Ling, and Tom Kwiatkowski.
  2019.
\newblock \href {https://doi.org/10.18653/v1/P19-1279} {Matching the blanks:
  Distributional similarity for relation learning}.
\newblock In \emph{Proceedings of the 57th Annual Meeting of the Association
  for Computational Linguistics}, pages 2895--2905, Florence, Italy.
  Association for Computational Linguistics.

\bibitem[{Biesialska et~al.(2020)Biesialska, Biesialska, and
  Costa-juss{\`a}}]{biesialska-etal-2020-continual}
Magdalena Biesialska, Katarzyna Biesialska, and Marta~R. Costa-juss{\`a}. 2020.
\newblock \href {https://doi.org/10.18653/v1/2020.coling-main.574} {Continual
  lifelong learning in natural language processing: A survey}.
\newblock In \emph{Proceedings of the 28th International Conference on
  Computational Linguistics}, pages 6523--6541, Barcelona, Spain (Online).
  International Committee on Computational Linguistics.

\bibitem[{Cui et~al.(2021)Cui, Yang, Yu, Hu, Cheng, Yi, and
  Xiao}]{cui-etal-2021-refining}
Li~Cui, Deqing Yang, Jiaxin Yu, Chengwei Hu, Jiayang Cheng, Jingjie Yi, and
  Yanghua Xiao. 2021.
\newblock \href {https://doi.org/10.18653/v1/2021.acl-long.20} {Refining sample
  embeddings with relation prototypes to enhance continual relation
  extraction}.
\newblock In \emph{Proceedings of the 59th Annual Meeting of the Association
  for Computational Linguistics and the 11th International Joint Conference on
  Natural Language Processing (Volume 1: Long Papers)}, pages 232--243, Online.
  Association for Computational Linguistics.

\bibitem[{Gormley et~al.(2015)Gormley, Yu, and
  Dredze}]{gormley-etal-2015-improved}
Matthew~R. Gormley, Mo~Yu, and Mark Dredze. 2015.
\newblock \href {https://doi.org/10.18653/v1/D15-1205} {Improved relation
  extraction with feature-rich compositional embedding models}.
\newblock In \emph{Proceedings of the 2015 Conference on Empirical Methods in
  Natural Language Processing}, pages 1774--1784, Lisbon, Portugal. Association
  for Computational Linguistics.

\bibitem[{Han et~al.(2020)Han, Dai, Gao, Lin, Liu, Li, Sun, and
  Zhou}]{han-etal-2020-continual}
Xu~Han, Yi~Dai, Tianyu Gao, Yankai Lin, Zhiyuan Liu, Peng Li, Maosong Sun, and
  Jie Zhou. 2020.
\newblock \href {https://doi.org/10.18653/v1/2020.acl-main.573} {Continual
  relation learning via episodic memory activation and reconsolidation}.
\newblock In \emph{Proceedings of the 58th Annual Meeting of the Association
  for Computational Linguistics}, pages 6429--6440, Online. Association for
  Computational Linguistics.

\bibitem[{Han et~al.(2018{\natexlab{a}})Han, Yu, Liu, Sun, and
  Li}]{han-etal-2018-hierarchical}
Xu~Han, Pengfei Yu, Zhiyuan Liu, Maosong Sun, and Peng Li. 2018{\natexlab{a}}.
\newblock \href {https://doi.org/10.18653/v1/D18-1247} {Hierarchical relation
  extraction with coarse-to-fine grained attention}.
\newblock In \emph{Proceedings of the 2018 Conference on Empirical Methods in
  Natural Language Processing}, pages 2236--2245, Brussels, Belgium.
  Association for Computational Linguistics.

\bibitem[{Han et~al.(2018{\natexlab{b}})Han, Zhu, Yu, Wang, Yao, Liu, and
  Sun}]{han-etal-2018-fewrel}
Xu~Han, Hao Zhu, Pengfei Yu, Ziyun Wang, Yuan Yao, Zhiyuan Liu, and Maosong
  Sun. 2018{\natexlab{b}}.
\newblock \href {https://doi.org/10.18653/v1/D18-1514} {{F}ew{R}el: A
  large-scale supervised few-shot relation classification dataset with
  state-of-the-art evaluation}.
\newblock In \emph{Proceedings of the 2018 Conference on Empirical Methods in
  Natural Language Processing}, pages 4803--4809, Brussels, Belgium.
  Association for Computational Linguistics.

\bibitem[{Khan et~al.(2019)Khan, Hayat, Zamir, Shen, and Shao}]{8953234}
Salman Khan, Munawar Hayat, Syed~Waqas Zamir, Jianbing Shen, and Ling Shao.
  2019.
\newblock \href {https://doi.org/10.1109/CVPR.2019.00019} {Striking the right
  balance with uncertainty}.
\newblock In \emph{2019 IEEE/CVF Conference on Computer Vision and Pattern
  Recognition (CVPR)}, pages 103--112.

\bibitem[{Kim and Kim(2020)}]{9081988}
Byungju Kim and Junmo Kim. 2020.
\newblock \href {https://doi.org/10.1109/ACCESS.2020.2991231} {Adjusting
  decision boundary for class imbalanced learning}.
\newblock \emph{IEEE Access}, 8:81674--81685.

\bibitem[{Kirkpatrick et~al.(2017)Kirkpatrick, Pascanu, Rabinowitz, Veness,
  Desjardins, Rusu, Milan, Quan, Ramalho, Grabska-Barwinska
  et~al.}]{regularization-3}
James Kirkpatrick, Razvan Pascanu, Neil Rabinowitz, Joel Veness, Guillaume
  Desjardins, Andrei~A Rusu, Kieran Milan, John Quan, Tiago Ramalho, Agnieszka
  Grabska-Barwinska, et~al. 2017.
\newblock \href {https://www.pnas.org/doi/abs/10.1073/pnas.1611835114}
  {Overcoming catastrophic forgetting in neural networks}.
\newblock \emph{Proceedings of the national academy of sciences},
  114(13):3521--3526.

\bibitem[{Li and Hoiem(2017)}]{regularization-1}
Zhizhong Li and Derek Hoiem. 2017.
\newblock \href {https://ieeexplore.ieee.org/abstract/document/8107520}
  {Learning without forgetting}.
\newblock \emph{IEEE transactions on pattern analysis and machine
  intelligence}, 40(12):2935--2947.

\bibitem[{Lin et~al.(2016)Lin, Shen, Liu, Luan, and Sun}]{lin-etal-2016-neural}
Yankai Lin, Shiqi Shen, Zhiyuan Liu, Huanbo Luan, and Maosong Sun. 2016.
\newblock \href {https://doi.org/10.18653/v1/P16-1200} {Neural relation
  extraction with selective attention over instances}.
\newblock In \emph{Proceedings of the 54th Annual Meeting of the Association
  for Computational Linguistics (Volume 1: Long Papers)}, pages 2124--2133,
  Berlin, Germany. Association for Computational Linguistics.

\bibitem[{Liu et~al.(2018)Liu, Zoph, Neumann, Shlens, Hua, Li, Fei-Fei, Yuille,
  Huang, and Murphy}]{isolation-4}
Chenxi Liu, Barret Zoph, Maxim Neumann, Jonathon Shlens, Wei Hua, Li-Jia Li,
  Li~Fei-Fei, Alan Yuille, Jonathan Huang, and Kevin Murphy. 2018.
\newblock \href
  {https://openaccess.thecvf.com/content_ECCV_2018/html/Chenxi_Liu_Progressive_Neural_Architecture_ECCV_2018_paper.html}
  {Progressive neural architecture search}.
\newblock In \emph{Proceedings of the European conference on computer vision
  (ECCV)}, pages 19--34.

\bibitem[{Liu et~al.(2019)Liu, Wang, Wu, Jiao, Wang, Xie, and
  Sun}]{liu2019reet}
Hongtao Liu, Peiyi Wang, Fangzhao Wu, Pengfei Jiao, Wenjun Wang, Xing Xie, and
  Yueheng Sun. 2019.
\newblock \href
  {https://link.springer.com/chapter/10.1007/978-3-030-32233-5_26} {Reet: Joint
  relation extraction and entity typing via multi-task learning}.
\newblock In \emph{CCF International Conference on Natural Language Processing
  and Chinese Computing}, pages 327--339. Springer.

\bibitem[{Lopez-Paz and Ranzato(2017)}]{replay-1}
David Lopez-Paz and Marc'Aurelio Ranzato. 2017.
\newblock \href
  {https://proceedings.neurips.cc/paper/2017/hash/f87522788a2be2d171666752f97ddebb-Abstract.html}
  {Gradient episodic memory for continual learning}.
\newblock \emph{Advances in neural information processing systems},
  30:6467--6476.

\bibitem[{Mallya et~al.(2018)Mallya, Davis, and Lazebnik}]{isolation-2}
Arun Mallya, Dillon Davis, and Svetlana Lazebnik. 2018.
\newblock \href
  {https://openaccess.thecvf.com/content_ECCV_2018/papers/Arun_Mallya_Piggyback_Adapting_a_ECCV_2018_paper.pdf}
  {Piggyback: Adapting a single network to multiple tasks by learning to mask
  weights}.
\newblock In \emph{Proceedings of the European Conference on Computer Vision
  (ECCV)}, pages 67--82.

\bibitem[{Mallya and Lazebnik(2018)}]{isolation-1}
Arun Mallya and Svetlana Lazebnik. 2018.
\newblock \href
  {https://openaccess.thecvf.com/content_cvpr_2018/papers/Mallya_PackNet_Adding_Multiple_CVPR_2018_paper.pdf}
  {Packnet: Adding multiple tasks to a single network by iterative pruning}.
\newblock In \emph{Proceedings of the IEEE conference on Computer Vision and
  Pattern Recognition}, pages 7765--7773.

\bibitem[{Mintz et~al.(2009)Mintz, Bills, Snow, and
  Jurafsky}]{mintz-etal-2009-distant}
Mike Mintz, Steven Bills, Rion Snow, and Daniel Jurafsky. 2009.
\newblock \href {https://aclanthology.org/P09-1113} {Distant supervision for
  relation extraction without labeled data}.
\newblock In \emph{Proceedings of the Joint Conference of the 47th Annual
  Meeting of the {ACL} and the 4th International Joint Conference on Natural
  Language Processing of the {AFNLP}}, pages 1003--1011, Suntec, Singapore.
  Association for Computational Linguistics.

\bibitem[{Miwa and Bansal(2016)}]{miwa-bansal-2016-end}
Makoto Miwa and Mohit Bansal. 2016.
\newblock \href {https://doi.org/10.18653/v1/P16-1105} {End-to-end relation
  extraction using {LSTM}s on sequences and tree structures}.
\newblock In \emph{Proceedings of the 54th Annual Meeting of the Association
  for Computational Linguistics (Volume 1: Long Papers)}, pages 1105--1116,
  Berlin, Germany. Association for Computational Linguistics.

\bibitem[{Riedel et~al.(2013)Riedel, Yao, McCallum, and
  Marlin}]{riedel-etal-2013-relation}
Sebastian Riedel, Limin Yao, Andrew McCallum, and Benjamin~M. Marlin. 2013.
\newblock \href {https://aclanthology.org/N13-1008} {Relation extraction with
  matrix factorization and universal schemas}.
\newblock In \emph{Proceedings of the 2013 Conference of the North {A}merican
  Chapter of the Association for Computational Linguistics: Human Language
  Technologies}, pages 74--84, Atlanta, Georgia. Association for Computational
  Linguistics.

\bibitem[{Seff et~al.(2017)Seff, Beatson, Suo, and Liu}]{replay-3}
Ari Seff, Alex Beatson, Daniel Suo, and Han Liu. 2017.
\newblock \href {https://arxiv.org/abs/1705.08395} {Continual learning in
  generative adversarial nets}.
\newblock \emph{arXiv preprint arXiv:1705.08395}.

\bibitem[{Serra et~al.(2018)Serra, Suris, Miron, and Karatzoglou}]{isolation-3}
Joan Serra, Didac Suris, Marius Miron, and Alexandros Karatzoglou. 2018.
\newblock \href {https://proceedings.mlr.press/v80/serra18a.html} {Overcoming
  catastrophic forgetting with hard attention to the task}.
\newblock In \emph{International Conference on Machine Learning}, pages
  4548--4557. PMLR.

\bibitem[{Shin et~al.(2017)Shin, Lee, Kim, and Kim}]{replay-2}
Hanul Shin, Jung~Kwon Lee, Jaehong Kim, and Jiwon Kim. 2017.
\newblock \href
  {https://proceedings.neurips.cc/paper/2017/file/0efbe98067c6c73dba1250d2beaa81f9-Paper.pdf}
  {Continual learning with deep generative replay}.
\newblock In \emph{Proceedings of the 31st International Conference on Neural
  Information Processing Systems}, pages 2994--3003.

\bibitem[{Wang et~al.(2019)Wang, Xiong, Yu, Guo, Chang, and
  Wang}]{wang-etal-2019-sentence}
Hong Wang, Wenhan Xiong, Mo~Yu, Xiaoxiao Guo, Shiyu Chang, and William~Yang
  Wang. 2019.
\newblock \href {https://doi.org/10.18653/v1/N19-1086} {Sentence embedding
  alignment for lifelong relation extraction}.
\newblock In \emph{Proceedings of the 2019 Conference of the North {A}merican
  Chapter of the Association for Computational Linguistics: Human Language
  Technologies, Volume 1 (Long and Short Papers)}, pages 796--806, Minneapolis,
  Minnesota. Association for Computational Linguistics.

\bibitem[{Wu et~al.(2021)Wu, Li, Li, Haffari, Qi, Zhu, and
  Xu}]{wu2021curriculum}
Tongtong Wu, Xuekai Li, Yuan-Fang Li, Gholamreza Haffari, Guilin Qi, Yujin Zhu,
  and Guoqiang Xu. 2021.
\newblock \href {https://arxiv.org/abs/2101.01926} {Curriculum-meta learning
  for order-robust continual relation extraction}.
\newblock In \emph{Proceedings of the AAAI Conference on Artificial
  Intelligence}, volume~35, pages 10363--10369.

\bibitem[{Xu et~al.(2015{\natexlab{a}})Xu, Mou, Li, Chen, Peng, and
  Jin}]{xu-etal-2015-classifying}
Yan Xu, Lili Mou, Ge~Li, Yunchuan Chen, Hao Peng, and Zhi Jin.
  2015{\natexlab{a}}.
\newblock \href {https://doi.org/10.18653/v1/D15-1206} {Classifying relations
  via long short term memory networks along shortest dependency paths}.
\newblock In \emph{Proceedings of the 2015 Conference on Empirical Methods in
  Natural Language Processing}, pages 1785--1794, Lisbon, Portugal. Association
  for Computational Linguistics.

\bibitem[{Xu et~al.(2015{\natexlab{b}})Xu, Mou, Li, Chen, Peng, and
  Jin}]{sup-re-xu-etal-2015-classifying}
Yan Xu, Lili Mou, Ge~Li, Yunchuan Chen, Hao Peng, and Zhi Jin.
  2015{\natexlab{b}}.
\newblock \href {https://doi.org/10.18653/v1/D15-1206} {Classifying relations
  via long short term memory networks along shortest dependency paths}.
\newblock In \emph{Proceedings of the 2015 Conference on Empirical Methods in
  Natural Language Processing}, pages 1785--1794, Lisbon, Portugal. Association
  for Computational Linguistics.

\bibitem[{Zelenko et~al.(2003)Zelenko, Aone, and
  Richardella}]{sup-re-zelenko2003kernel}
Dmitry Zelenko, Chinatsu Aone, and Anthony Richardella. 2003.
\newblock \href {https://www.jmlr.org/papers/volume3/tmp/zelenko03a.pdf}
  {Kernel methods for relation extraction}.
\newblock \emph{Journal of machine learning research}, 3(Feb):1083--1106.

\bibitem[{Zeng et~al.(2015)Zeng, Liu, Chen, and Zhao}]{zeng-etal-2015-distant}
Daojian Zeng, Kang Liu, Yubo Chen, and Jun Zhao. 2015.
\newblock \href {https://doi.org/10.18653/v1/D15-1203} {Distant supervision for
  relation extraction via piecewise convolutional neural networks}.
\newblock In \emph{Proceedings of the 2015 Conference on Empirical Methods in
  Natural Language Processing}, pages 1753--1762, Lisbon, Portugal. Association
  for Computational Linguistics.

\bibitem[{Zeng et~al.(2014)Zeng, Liu, Lai, Zhou, and
  Zhao}]{zeng-etal-2014-relation}
Daojian Zeng, Kang Liu, Siwei Lai, Guangyou Zhou, and Jun Zhao. 2014.
\newblock \href {https://aclanthology.org/C14-1220} {Relation classification
  via convolutional deep neural network}.
\newblock In \emph{Proceedings of {COLING} 2014, the 25th International
  Conference on Computational Linguistics: Technical Papers}, pages 2335--2344,
  Dublin, Ireland. Dublin City University and Association for Computational
  Linguistics.

\bibitem[{Zhang et~al.(2018)Zhang, Qi, and Manning}]{zhang-etal-2018-graph}
Yuhao Zhang, Peng Qi, and Christopher~D. Manning. 2018.
\newblock \href {https://doi.org/10.18653/v1/D18-1244} {Graph convolution over
  pruned dependency trees improves relation extraction}.
\newblock In \emph{Proceedings of the 2018 Conference on Empirical Methods in
  Natural Language Processing}, pages 2205--2215, Brussels, Belgium.
  Association for Computational Linguistics.

\bibitem[{Zhang et~al.(2017)Zhang, Zhong, Chen, Angeli, and
  Manning}]{zhang-etal-2017-position}
Yuhao Zhang, Victor Zhong, Danqi Chen, Gabor Angeli, and Christopher~D.
  Manning. 2017.
\newblock \href {https://doi.org/10.18653/v1/D17-1004} {Position-aware
  attention and supervised data improve slot filling}.
\newblock In \emph{Proceedings of the 2017 Conference on Empirical Methods in
  Natural Language Processing}, pages 35--45, Copenhagen, Denmark. Association
  for Computational Linguistics.

\bibitem[{Zhao et~al.(2022)Zhao, Xu, Yang, and Gao}]{zhao2022consistent}
Kang Zhao, Hua Xu, Jiangong Yang, and Kai Gao. 2022.
\newblock \href {https://arxiv.org/abs/2203.02721} {Consistent representation
  learning for continual relation extraction}.
\newblock \emph{arXiv preprint arXiv:2203.02721}.

\bibitem[{Zhou et~al.(2005)Zhou, Su, Zhang, and
  Zhang}]{sup-re-zhou-etal-2005-exploring}
GuoDong Zhou, Jian Su, Jie Zhang, and Min Zhang. 2005.
\newblock \href {https://doi.org/10.3115/1219840.1219893} {Exploring various
  knowledge in relation extraction}.
\newblock In \emph{Proceedings of the 43rd Annual Meeting of the Association
  for Computational Linguistics ({ACL}{'}05)}, pages 427--434, Ann Arbor,
  Michigan. Association for Computational Linguistics.

\end{thebibliography}
\bibliographystyle{acl_natbib}

\appendix

\clearpage
\newpage
\section{Baselines}
\label{app:baseline}
We compared \modelname with following baselines:
\begin{itemize}
\item \textbf{EA-EMR} \cite{wang-etal-2019-sentence}, which uses an embedding alignment mechanism to maintain memory to alleviate the catastrophic forgetting problem.
\item \textbf{CML} \cite{wu2021curriculum}, which proposes a curriculum-meta learning method to alleviate catastrophic forgetting while maintaining order-robust.
\item \textbf{EMAR} \cite{han-etal-2020-continual}, which introduces memory activation and reconsolidation mechanism for continual relation extraction.
\item \textbf{RP-CRE} \cite{cui-etal-2021-refining}, which refines sample embeddings with relation prototypes to enhance continual relation extraction.
\item \textbf{CRL} \cite{zhao2022consistent}, which proposes a consistent representation learning method that utilizes contrastive replay and knowledge distillation to alleviate catastrophic forgetting.
\end{itemize}

\label{app:ablation}
\begin{table*}[t]
\centering
\scalebox{0.95}{
    \begin{tabular}{lcccc}
    \toprule
    \textbf{Ablation Methods} & \textbf{\textsc{Stage} 1} & \textbf{\textsc{Stage} 2} & \textbf{\fewrel} & \textbf{\tacred}  \\
    \midrule
    \modelname & $\mathcal{D}_k$ & $\mathcal{M}_k$  & 84.3 & 77.7   \\
    \midrule
    A1 (\textit{remove} BT) & $\mathcal{D}_k$ & $\mathcal{M}_{k-1} \cup $ $\mathcal{D}_k$ & 75.8 & 71.2 \\
    A2 (\textit{remove} FA)  & $\mathcal{M}_{k-1} \cup \mathcal{D}_k$ & $\mathcal{M}_k$  & 81.1 & 74.9 \\
    A3 (\textit{remove} FA and BT) & $\emptyset$ & $\mathcal{M}_{k-1} \cup \mathcal{D}_k$  & 75.6 & 71.2 \\
    A4  & $\emptyset$ & $\mathcal{M}_{k-1} \cup \mathcal{M}_{k-1} \cup \dots \cup \mathcal{M}_{k-1} \cup $ $\mathcal{D}_k$  & 73.1 & 70.1 \\
    A5  & $\mathcal{D}_k$ &$\mathcal{M}_{k-1} \cup \mathcal{M}_{k-1} \cup \dots \cup \mathcal{M}_{k-1} \cup $ $\mathcal{D}_k$  & 78.1 & 73.8 \\
    \bottomrule
    \end{tabular}
}
\caption{Performances of models with different training data at two learning stages. $\mathcal{D}_k$ is the data of new relations, $\mathcal{M}_{k-1}$ is the memory data of previous relations and $\mathcal{M}_{k}$ is the memory data of previous and new relations.}
\label{tab:app-ablation}
\end{table*}

\section{Influence of Memory Size}
\begin{figure}[t]
    \centering
    \includegraphics[width=\linewidth]{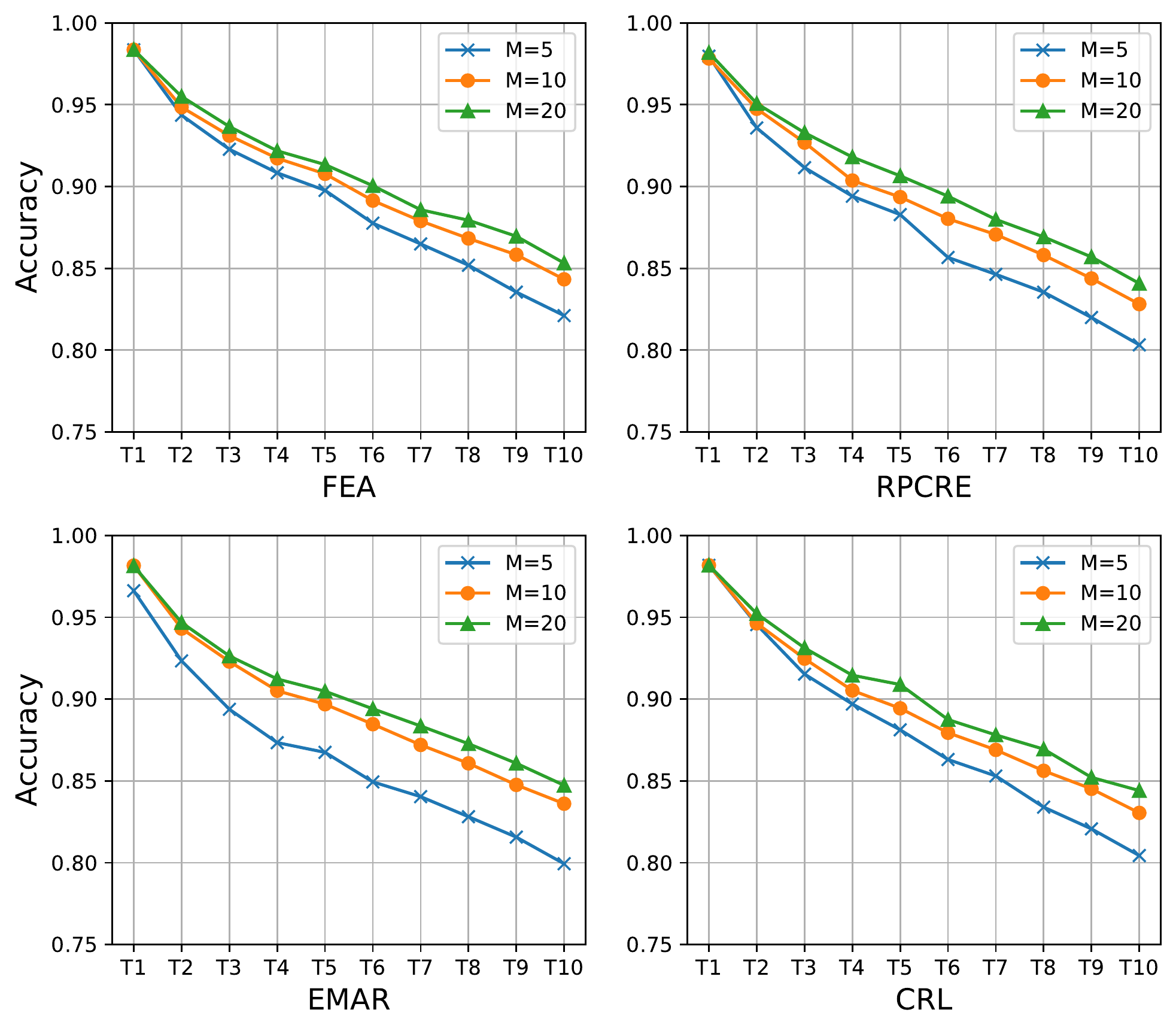}
    \caption{The results of different models with different memory sizes on \fewrel. The X-axis is the task stage and Y-axis is the accuracy on the test set of all observed relations. We report the average of $5$ different runs.}
    \label{fig:memory-fewrel}
\end{figure}
\label{app:mem}

\begin{figure}[t]
    \centering
    \includegraphics[width=\linewidth]{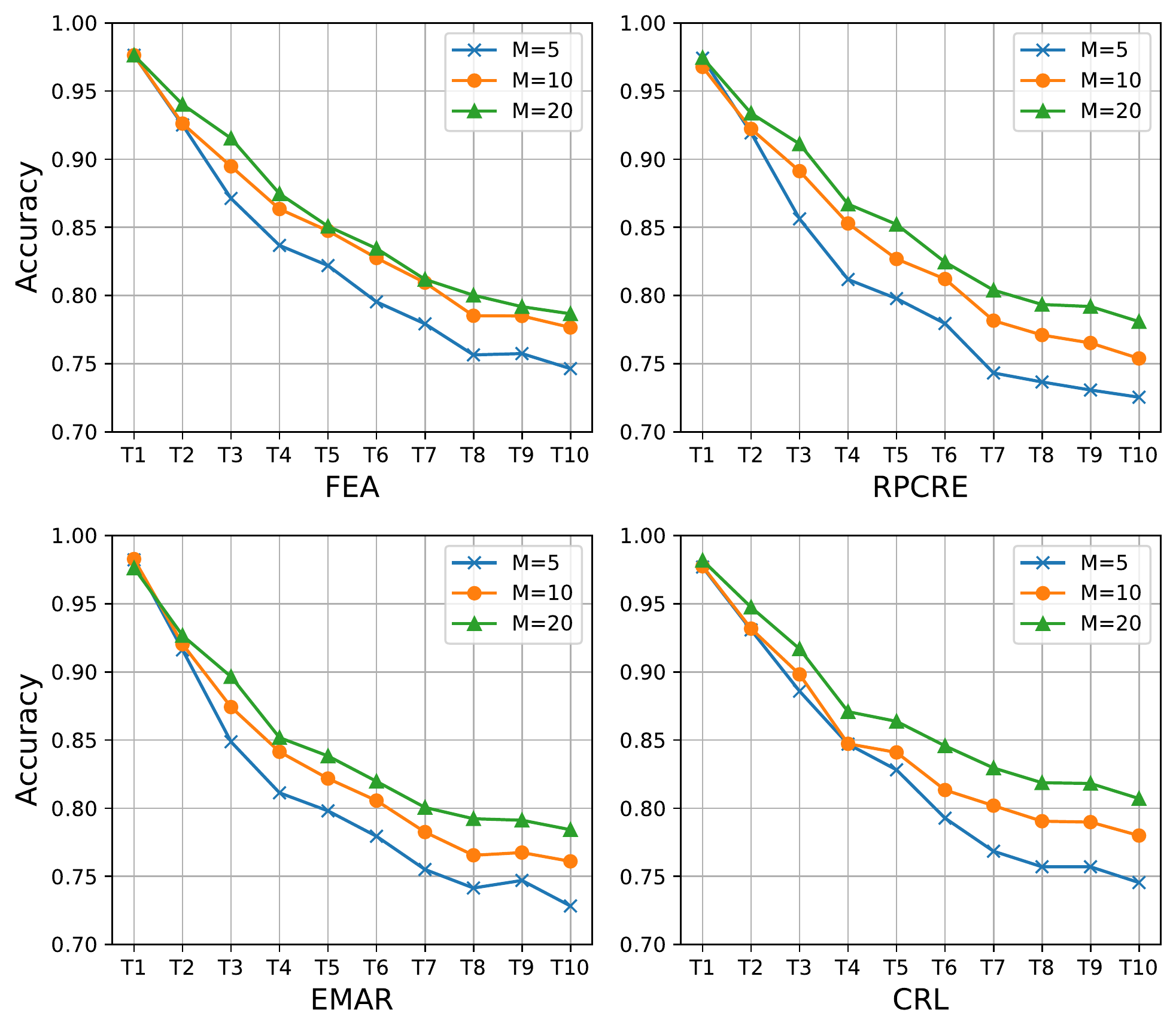}
    \caption{The results of different models with different memory sizes on \tacred.  We report the average of $5$ different runs.}
    \label{fig:memory-tacred}
\end{figure}
\label{app:mem}

For the rehearsal-based CRE models, memory size (the number of stored instances for each relation) is a key factor for the model performance.
Therefore, in this section, we also study the performance of \modelname with different memory sizes.
As shown in Figure \ref{fig:memory-fewrel} and Figure \ref{fig:memory-tacred}:
\textbf{1)} As the size of the memory decreases, the performances of all models drop, which shows the importance of the memory size for CRE models.
\textbf{2)} In \fewrel, \modelname outperforms three baselines with different memory sizes. 
In addition, when decreasing the memory size, the performance gap between \modelname and baselines tends to be larger.
For example, on the final task, \modelname outperforms RPCRE $2.0$, $1.5$ and $1.1$ accuracy on memory size $5$, $10$ and $20$, respectively.
\textbf{3)} In \tacred, \modelname outperforms RPCRE and EMAR with three different memory sizes, especially when memory sizes are $5$ and $10$. 
CRL outperforms \modelname when memory size is $20$. However, when memory size is $10$ and $5$, \modelname achieves comparable results with CRL.
These results show \modelname has more obvious advantages when the memory size is small.
(4) By comparing the performance gap of $T10$ between $M=5$ and $M=20$, we can also find \modelname performs more stable than all baselines.

\section{Model Efficiency}
\label{app:me}
\begin{table}[t]
\centering
\small
\begin{tabular}{lcccc}
\toprule
\multirow{2}{*}{\textbf{Models}} & \multicolumn{2}{c}{\textbf{\fewrel}} & \multicolumn{2}{c}{\textbf{\tacred}}  \\
\cmidrule(r){2-3} \cmidrule(r){4-5}
& \textbf{Accuracy} & \textbf{Time} & \textbf{Accuracy} & \textbf{Time} \\
\midrule
\modelname  & 84.3 & 173.7s  & 77.7 & 53.8s   \\
\midrule
RPCRE & 82.8 & 347.2s & 75.3 & 89.8s\\
EMAR & 83.6 & 297.3s & 76.1 & 70.8s \\
CRL   & 83.1 & 256.2s &  78.0 & 67.7s \\

\bottomrule
\end{tabular}
\caption{Performances and training time of different models on two benchmarks. Compared with baselines, \modelname takes less training time to update the model.}
\label{tab:efficiency}
\end{table}
In real life, CRE models need to learn new relations constantly, thus we should also consider the learning efficiency of models.
In this section, we compared the learning time (average training time per task) of \modelname and our baselines on two benchmarks.
As shown in Table \ref{tab:efficiency}, with exactly the same hardware setting, compared with all strong baselines, \modelname has a better learning efficiency.

\section{More Detail of Ablation Study}
\label{app:ablation}
In Table \ref{tab:app-ablation}, We explore $5$ different kinds of ablation methods in total.
$\mathcal{M}_{k-1} \cup \mathcal{M}_{k-1} \cup \dots \cup \mathcal{M}_{k-1} \cup $ $\mathcal{D}_k$ denotes the up-sampling memory data where the number of memory data for each previously learned relation is the same as that of its origin training data.
As is shown:
\textbf{1)} A3 and A4 have poor performances on two benchmarks, showing that it is important to propose a two-stage learning CRE model.
\textbf{2)}  A5 outperforms A4, showing that it is important to learn new relations at stage 1.
\modelname outperform A2 shows that stage 1 should have ONLY new data.
\textbf{3)} \modelname and A2 outperform all other methods showing that it is important to ensure the data imbalance at stage 2.

\section{More Detail of Error Analysis}
\label{app:error}

To better explore the effectiveness of BT and FA, we conduct error analysis for \modelname, ``BT: \textit{remove} BT'' and ``\textit{remove} FA'' on \fewrel in in Table \ref{tab:error1} and Table \ref{tab:error2}.
From the results, we can find:
\textbf{1)} The model makes a lot of \textit{latter error}. 
\textbf{2)} The model wrongly predicts many instances as the relations appearing at the last task (i.e., [10]). 
\textbf{3)} The model is unable to distinguish similarity relations, and easily predicts former-occurring relation instances as the later-occurring similar relation. For example, ``followed by (P156, appear at  task [1])'' to ``follow (P155, appears at  task [8])'', ``mother'' (P25, appears at task [4]) to ``father'' (P22, appears at  task [10]), ``tributary (P974, appears at task [7])'' to ``mouth of the watercourse (P403, appears at task [9])''.
(4) Removing either BT or FA will significantly increase the \textit{latter error}, and thus aggravate the catastrophic forgetting.

\section{Effectiveness of BT}
\label{app:bt}
\begin{figure*}[th]
    \centering
    \includegraphics[width=\linewidth]{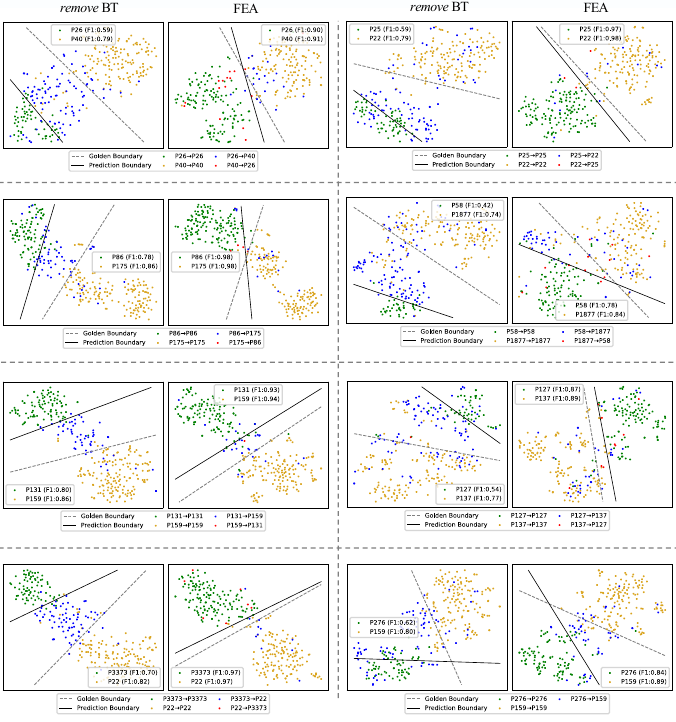}
    \caption{T-SNE of instances belonging to different similar relations pairs with server \textit{latter error}. BT helps build a more balanced decision boundary between old and new relations, and thus reduces the \textit{latter error}.}
    \label{fig:tsne-app-bt}
\end{figure*}

We draw several relation pairs with severe \textit{latter error} to show the effectiveness of BT in Figure \ref{fig:tsne-app-bt}.

\section{Effectiveness of FA}
\label{app:FA}
\begin{figure*}[th]
    \centering
    \includegraphics[width=0.95\linewidth]{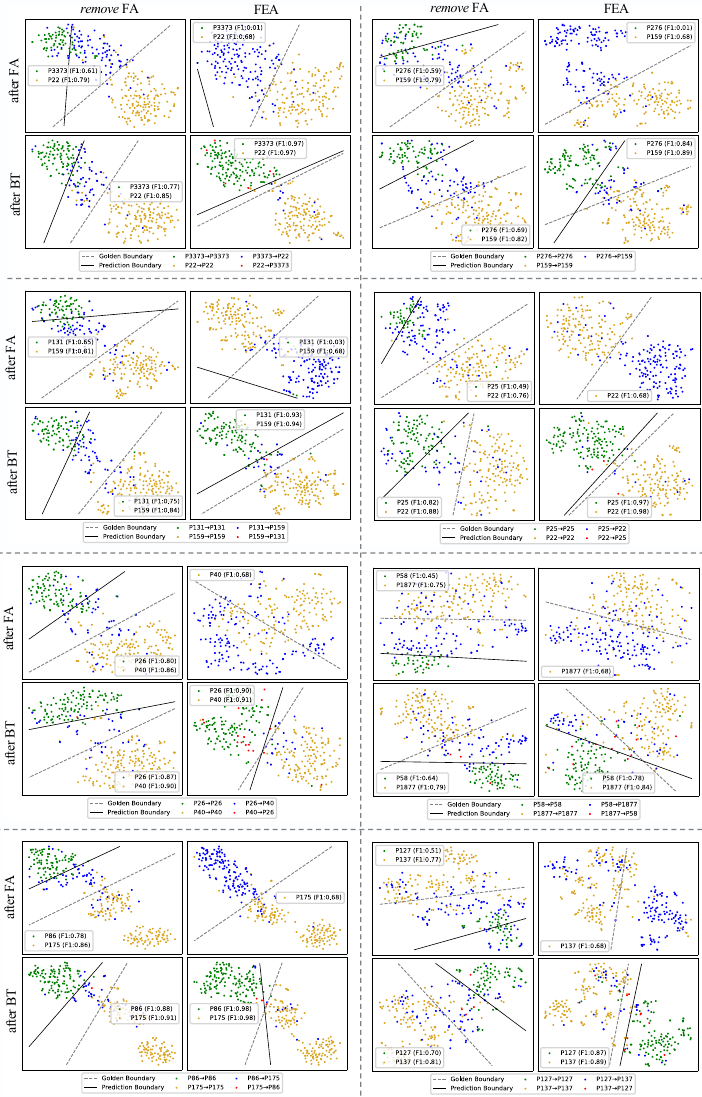}
    \caption{T-SNE of instances belonging to different similar relations pairs with server \textit{latter error}s. FA preserves the potential of memory instances for BT, and thus reduces the \textit{latter error}.}
    \label{fig:tsne-app}
\end{figure*}

We draw several relation pairs with severe \textit{latter error} to show the effectiveness of FA in Figure \ref{fig:tsne-app}.

\begin{table*}[t]
    \centering
    \scalebox{0.67}{
    \begin{tabular}{l|ccc|ccc|ccc}
    \toprule
      \multirow{2}{*}{Relations}  & \multicolumn{3}{c}{\modelname} & \multicolumn{3}{c}{\textit{remove} BT} & \multicolumn{3}{c}{\textit{remove} FA} \\
        \cmidrule(r){2-4} \cmidrule(r){5-7}  \cmidrule(r){8-10} 
      & Acc. & TOP1 & TOP2 & Acc. & TOP1 & TOP2  & Acc. & TOP1 & TOP2 \\
      \midrule
P156/[1] & \color{red}{0.52} & \color{red}{P155/[8]/0.31} & \color{red}{P176/[10]/0.02} & \color{red}{0.36} & \color{red}{P155/[8]/0.31} & \color{red}{P31/[10]/0.15} & \color{red}{0.43} & \color{red}{P155/[8]/0.27} & \color{red}{P31/[10]/0.09}\\
P84/[1] & 0.94 & P127/[5]/0.03 & P6/[3]/0.01 & 0.91 & P176/[10]/0.04 & P127/[5]/0.02 & 0.93 & P31/[10]/0.01 & P1877/[10]/0.01\\
P39/[1] & 0.94 & P106/[4]/0.02 & P410/[1]/0.01 & 0.88 & P31/[10]/0.04 & P106/[4]/0.04 & 0.92 & P106/[4]/0.04 & P31/[10]/0.01\\
P276/[1] & \color{red}{0.52} & \color{red}{P159/[10]/0.21} & \color{red}{P551/[7]/0.04} & \color{red}{0.34} & \color{red}{P159/[10]/0.52} & \color{red}{P31/[10]/0.04} & \color{red}{0.46} & \color{red}{P159/[10]/0.37} & \color{red}{P706/[5]/0.04}\\
P410/[1] & 0.99 & P39/[1]/0.01 & P106/[4]/0.01 & 0.98 & P241/[1]/0.01 & P31/[10]/0.01 & 0.99 & P106/[4]/0.01 & P26/[8]/0.00\\
P241/[1] & 0.91 & P137/[7]/0.05 & P27/[8]/0.01 & 0.88 & P137/[7]/0.04 & P407/[10]/0.04 & 0.81 & P137/[7]/0.11 & P27/[8]/0.03\\
P177/[1] & 1.00 & P137/[7]/0.00 & P937/[7]/0.00 & 0.98 & P4552/[6]/0.01 & P206/[6]/0.01 & 0.99 & P206/[6]/0.01 & P3450/[7]/0.00\\
P264/[1] & 0.95 & P175/[9]/0.03 & P463/[5]/0.01 & 0.92 & P175/[9]/0.02 & P176/[10]/0.01 & 0.90 & P175/[9]/0.04 & P750/[4]/0.03\\
\midrule
P412/[2] & 0.99 & P1303/[8]/0.01 & P178/[7]/0.00 & 0.97 & P1303/[8]/0.03 & P178/[7]/0.00 & 0.99 & P1303/[8]/0.01 & P178/[7]/0.00\\
P361/[2] & \color{red}{0.27} & \color{red}{P31/[10]/0.11} & \color{red}{P131/[2]/0.07} & \color{red}{0.21} & \color{red}{P31/[10]/0.31} & \color{red}{P1344/[10]/0.08} & \color{red}{0.22} & \color{red}{P31/[10]/0.21} & \color{red}{P527/[8]/0.06}\\
P1923/[2] & 0.95 & P1346/[4]/0.03 & P710/[2]/0.01 & 0.87 & P1346/[4]/0.07 & P159/[10]/0.01 & 0.88 & P1346/[4]/0.07 & P710/[2]/0.04\\
P123/[2] & \color{red}{0.59} & \color{red}{P178/[7]/0.34} & \color{red}{P750/[4]/0.02} & \color{red}{0.45} & \color{red}{P178/[7]/0.37} & \color{red}{P176/[10]/0.06} & \color{red}{0.52} & \color{red}{P178/[7]/0.34} & \color{red}{P176/[10]/0.07}\\
P118/[2] & 0.96 & P641/[10]/0.02 & P1344/[10]/0.01 & 0.88 & P1344/[10]/0.06 & P641/[10]/0.04 & 0.93 & P1344/[10]/0.04 & P641/[10]/0.03\\
P131/[2] & \color{red}{0.66} & \color{red}{P159/[10]/0.09} & \color{red}{P706/[5]/0.09} & \color{red}{0.44} & \color{red}{P159/[10]/0.31} & \color{red}{P31/[10]/0.08} & \color{red}{0.46} & \color{red}{P159/[10]/0.21} & \color{red}{P17/[6]/0.06}\\
P710/[2] & 0.72 & P1346/[4]/0.06 & P1923/[2]/0.06 & \color{red}{0.66} & \color{red}{P1346/[4]/0.08} & \color{red}{P1877/[10]/0.06} & 0.79 & P1346/[4]/0.05 & P991/[4]/0.04\\
P355/[2] & \color{red}{0.69} & \color{red}{P527/[8]/0.11} & \color{red}{P127/[5]/0.03} & \color{red}{0.56} & \color{red}{P31/[10]/0.18} & \color{red}{P159/[10]/0.15} & \color{red}{0.60} & \color{red}{P159/[10]/0.10} & \color{red}{P127/[5]/0.07}\\
\midrule
P6/[3] & 0.99 & P991/[4]/0.01 & P355/[2]/0.01 & 0.98 & P991/[4]/0.01 & P22/[10]/0.01 & 0.99 & P1877/[10]/0.01 & P991/[4]/0.01\\
P400/[3] & 0.91 & P306/[9]/0.09 & P31/[10]/0.01 & 0.84 & P306/[9]/0.11 & P31/[10]/0.04 & 0.82 & P306/[9]/0.16 & P361/[2]/0.01\\
P101/[3] & 0.79 & P135/[6]/0.04 & P31/[10]/0.03 & \color{red}{0.61} & \color{red}{P31/[10]/0.15} & \color{red}{P106/[4]/0.07} & \color{red}{0.68} & \color{red}{P106/[4]/0.10} & \color{red}{P31/[10]/0.06}\\
P140/[3] & 0.95 & P407/[10]/0.01 & P101/[3]/0.01 & 0.88 & P31/[10]/0.06 & P407/[10]/0.03 & 0.93 & P407/[10]/0.02 & P31/[10]/0.01\\
P2094/[3] & 1.00 & P137/[7]/0.00 & P937/[7]/0.00 & 0.99 & P1344/[10]/0.01 & P937/[7]/0.00 & 0.99 & P1344/[10]/0.01 & P937/[7]/0.00\\
P364/[3] & 0.86 & P407/[10]/0.13 & P27/[8]/0.01 & \color{red}{0.67} & \color{red}{P407/[10]/0.33} & \color{red}{P3450/[7]/0.00} & \color{red}{0.68} & \color{red}{P407/[10]/0.31} & \color{red}{P495/[4]/0.01}\\
P150/[3] & 0.94 & P1001/[6]/0.02 & P551/[7]/0.01 & 0.89 & P159/[10]/0.08 & P31/[10]/0.01 & 0.86 & P159/[10]/0.09 & P527/[8]/0.01\\
P466/[3] & 0.92 & P127/[5]/0.03 & P140/[3]/0.01 & 0.76 & P159/[10]/0.09 & P137/[7]/0.04 & 0.88 & P159/[10]/0.04 & P127/[5]/0.03\\
\midrule
P449/[4] & 0.98 & P750/[4]/0.01 & P1408/[5]/0.01 & 0.97 & P750/[4]/0.02 & P31/[10]/0.01 & 0.94 & P750/[4]/0.04 & P137/[7]/0.02\\
P674/[4] & 0.81 & P1877/[10]/0.06 & P527/[8]/0.04 & \color{red}{0.69} & \color{red}{P1877/[10]/0.25} & \color{red}{P527/[8]/0.03} & 0.78 & P1877/[10]/0.14 & P527/[8]/0.02\\
P991/[4] & 0.97 & P6/[3]/0.02 & P102/[9]/0.01 & 0.99 & P102/[9]/0.01 & P137/[7]/0.00 & 0.98 & P6/[3]/0.01 & P102/[9]/0.01\\
P495/[4] & 0.72 & P407/[10]/0.11 & P27/[8]/0.06 & \color{red}{0.31} & \color{red}{P407/[10]/0.46} & \color{red}{P27/[8]/0.08} & \color{red}{0.56} & \color{red}{P407/[10]/0.17} & \color{red}{P27/[8]/0.09}\\
P1346/[4] & 0.79 & P1923/[2]/0.14 & P710/[2]/0.05 & 0.83 & P1923/[2]/0.09 & P1877/[10]/0.04 & 0.82 & P1923/[2]/0.11 & P710/[2]/0.04\\
P750/[4] & 0.89 & P449/[4]/0.03 & P178/[7]/0.02 & 0.89 & P178/[7]/0.04 & P176/[10]/0.04 & 0.91 & P176/[10]/0.02 & P178/[7]/0.01\\
P106/[4] & \color{red}{0.69} & \color{red}{P641/[10]/0.13} & \color{red}{P101/[3]/0.11} & \color{red}{0.69} & \color{red}{P641/[10]/0.13} & \color{red}{P1303/[8]/0.07} & 0.79 & P641/[10]/0.11 & P1303/[8]/0.06\\
P25/[4] & 0.91 & P22/[10]/0.05 & P26/[8]/0.02 & \color{red}{0.39} & \color{red}{P22/[10]/0.59} & \color{red}{P26/[8]/0.03} & \color{red}{0.69} & \color{red}{P22/[10]/0.30} & \color{red}{P26/[8]/0.01}\\
\midrule
P706/[5] & \color{red}{0.67} & \color{red}{P206/[6]/0.12} & \color{red}{P4552/[6]/0.09} & \color{red}{0.59} & \color{red}{P206/[6]/0.11} & \color{red}{P31/[10]/0.11} & \color{red}{0.54} & \color{red}{P206/[6]/0.13} & \color{red}{P4552/[6]/0.10}\\
P127/[5] & \color{red}{0.51} & \color{red}{P137/[7]/0.17} & \color{red}{P176/[10]/0.09} & \color{red}{0.29} & \color{red}{P176/[10]/0.30} & \color{red}{P137/[7]/0.19} & \color{red}{0.36} & \color{red}{P137/[7]/0.21} & \color{red}{P176/[10]/0.20}\\
P413/[5] & 1.00 & P137/[7]/0.00 & P937/[7]/0.00 & 1.00 & P137/[7]/0.00 & P937/[7]/0.00 & 1.00 & P137/[7]/0.00 & P937/[7]/0.00\\
P463/[5] & 0.81 & P102/[9]/0.01 & P175/[9]/0.01 & \color{red}{0.64} & \color{red}{P31/[10]/0.16} & \color{red}{P175/[9]/0.05} & \color{red}{0.70} & \color{red}{P31/[10]/0.09} & \color{red}{P175/[9]/0.04}\\
P1408/[5] & 0.99 & P740/[5]/0.01 & P3450/[7]/0.00 & 0.91 & P159/[10]/0.09 & P407/[10]/0.01 & 0.98 & P159/[10]/0.01 & P17/[6]/0.01\\
P59/[5] & 0.99 & P22/[10]/0.01 & P137/[7]/0.00 & 0.95 & P31/[10]/0.04 & P22/[10]/0.01 & 0.99 & P460/[9]/0.01 & P31/[10]/0.01\\
P86/[5] & 0.74 & P58/[8]/0.10 & P175/[9]/0.06 & \color{red}{0.29} & \color{red}{P1877/[10]/0.51} & \color{red}{P175/[9]/0.15} & \color{red}{0.70} & \color{red}{P175/[9]/0.14} & \color{red}{P1877/[10]/0.12}\\
P740/[5] & 0.77 & P159/[10]/0.18 & P937/[7]/0.02 & \color{red}{0.67} & \color{red}{P159/[10]/0.32} & \color{red}{P27/[8]/0.01} & \color{red}{0.66} & \color{red}{P159/[10]/0.29} & \color{red}{P937/[7]/0.01}\\

        \bottomrule
    \end{tabular}
    }
    \caption{Error analysis of relations from task $1$ to task $5$ on \fewrel (relations from task $6$ to task $10$ are shown in Table \ref{tab:error2} due to the space limited.). ``Acc.'' denotes the accuracy. ``TOP1'' and ``TOP2'' denote the first and second confusing relations for each relation, respectively (model wrongly predicts instances as these relations). ``[*]'' denotes the task id that the relation appears. We highlight the results with $Acc. \leq 0.7$.
    From the results, we can find that:
    \textbf{1)} The model makes a lot of \textit{latter error}. 
    \textbf{2)} The model wrongly predicts many instances as the relations appearing at the last task (i.e., [10]). 
    \textbf{3)} The model is unable to distinguish similarity relations, and easily predicts former-occurring relation instances as the later-occurring similar relation. For example, ``followed by (P156, appear at  task [1])'' to ``follow (P155, appears at  task [8])'', ``mother'' (P25, appears at task [4]) to ``father'' (P22, appears at  task [10]), ``tributary (P974, appears at task [7])'' to ``mouth of the watercourse (P403, appears at task [9])'' and so on.
    (4) Removing either BT or FA will significantly increase the \textit{latter error}, and thus aggravates the catastrophic forgetting.
}
    \label{tab:error1}
\end{table*}

\begin{table*}[t]
    \centering
    \scalebox{0.65}{
    \begin{tabular}{l|ccc|ccc|ccc}
    \toprule
      \multirow{2}{*}{Relations}  & \multicolumn{3}{c}{\modelname} & \multicolumn{3}{c}{BT: \textit{remove} BT} & \multicolumn{3}{c}{\textit{remove} FA} \\
        \cmidrule(r){2-4} \cmidrule(r){5-7}  \cmidrule(r){8-10} 
      & Acc. & TOP1 & TOP2 & Acc. & TOP1 & TOP2  & Acc. & TOP1 & TOP2 \\
      \midrule

P206/[6] & 0.90 & P706/[5]/0.03 & P403/[9]/0.03 & 0.89 & P150/[3]/0.04 & P403/[9]/0.03 & 0.93 & P403/[9]/0.04 & P706/[5]/0.01\\
P17/[6] & 0.71 & P495/[4]/0.09 & {P407/[10]/0.07} & \color{red}{0.64} & \color{red}{P407/[10]/0.16} & \color{red}{P159/[10]/0.08} & 0.72 & P407/[10]/0.09 & P1001/[6]/0.07\\
P136/[6] & 0.87 & P31/[10]/0.05 & P641/[10]/0.02 & 0.76 & P31/[10]/0.15 & P641/[10]/0.03 & 0.82 & P31/[10]/0.09 & P641/[10]/0.03\\
P800/[6] & 0.94 & P176/[10]/0.01 & P551/[7]/0.01 & 0.89 & P31/[10]/0.06 & P176/[10]/0.01 & 0.93 & P86/[5]/0.01 & P31/[10]/0.01\\
P4552/[6] & 0.91 & P706/[5]/0.06 & P31/[10]/0.01 & 0.91 & P31/[10]/0.06 & P706/[5]/0.02 & 0.96 & P706/[5]/0.02 & P31/[10]/0.01\\
P1001/[6] & 0.81 & P495/[4]/0.06 & P17/[6]/0.04 & \color{red}{0.64} & \color{red}{P159/[10]/0.16} & \color{red}{P31/[10]/0.09} & 0.84 & P159/[10]/0.07 & P31/[10]/0.03\\
P931/[6] & 0.96 & P159/[10]/0.03 & P706/[5]/0.01 & 0.94 & P159/[10]/0.06 & P137/[7]/0.01 & 0.94 & P159/[10]/0.06 & P137/[7]/0.00\\
P135/[6] & 0.88 & P463/[5]/0.03 & P937/[7]/0.01 & 0.85 & P31/[10]/0.06 & P1344/[10]/0.01 & 0.88 & P31/[10]/0.02 & P101/[3]/0.02\\
\midrule
P3373/[7] & \color{red}{0.68} & \color{red}{P22/[10]/0.16} & \color{red}{P40/[9]/0.11} & \color{red}{0.54} & \color{red}{P22/[10]/0.38} & \color{red}{P40/[9]/0.04} & 0.77 & P22/[10]/0.09 & P40/[9]/0.09\\
P551/[7] & \color{red}{0.69} & \color{red}{P937/[7]/0.20} & \color{red}{P27/[8]/0.04} & \color{red}{0.56} & \color{red}{P159/[10]/0.18} & \color{red}{P937/[7]/0.14} & \color{red}{0.63} & \color{red}{P937/[7]/0.21} & \color{red}{P159/[10]/0.07}\\
P974/[7] & 0.90 & P403/[9]/0.10 & P3450/[7]/0.00 & \color{red}{0.60} & \color{red}{P403/[9]/0.39} & \color{red}{P131/[2]/0.01} & \color{red}{0.54} & \color{red}{P403/[9]/0.45} & \color{red}{P155/[8]/0.01}\\
P921/[7] & 0.75 & P31/[10]/0.05 & P641/[10]/0.04 & \color{red}{0.51} & \color{red}{P31/[10]/0.29} & \color{red}{P1877/[10]/0.06} & \color{red}{0.67} & \color{red}{P31/[10]/0.13} & \color{red}{P1877/[10]/0.04}\\
P3450/[7] & 0.99 & P641/[10]/0.01 & P937/[7]/0.00 & 0.93 & P31/[10]/0.04 & P1344/[10]/0.01 & 0.93 & P31/[10]/0.03 & P641/[10]/0.02\\
P137/[7] & 0.74 & P176/[10]/0.09 & P127/[5]/0.07 & 0.71 & {P176/[10]/0.12} & {P31/[10]/0.04} & 0.74 & P176/[10]/0.13 & P159/[10]/0.04\\
P937/[7] & 0.83 & P551/[7]/0.06 & P159/[10]/0.03 & 0.74 & P159/[10]/0.12 & P551/[7]/0.09 & 0.77 & P551/[7]/0.07 & P159/[10]/0.06\\
P178/[7] & 0.81 & P123/[2]/0.06 & P176/[10]/0.04 & \color{red}{0.69} & \color{red}{P176/[10]/0.20} & \color{red}{P31/[10]/0.04} & 0.74 & P176/[10]/0.13 & P306/[9]/0.03\\
\midrule
P26/[8] & \color{red}{0.69} & \color{red}{P40/[9]/0.11} & \color{red}{P22/[10]/0.10} & \color{red}{0.54} & \color{red}{P22/[10]/0.36} & \color{red}{P40/[9]/0.05} & \color{red}{0.67} & \color{red}{P22/[10]/0.16} & \color{red}{P40/[9]/0.09}\\
P1303/[8] & 1.00 & P137/[7]/0.00 & P937/[7]/0.00 & 0.99 & P1344/[10]/0.01 & P137/[7]/0.00 & 1.00 & P137/[7]/0.00 & P937/[7]/0.00\\
P1435/[8] & 1.00 & P137/[7]/0.00 & P3450/[7]/0.00 & 1.00 & P137/[7]/0.00 & P3450/[7]/0.00 & 1.00 & P137/[7]/0.00 & P3450/[7]/0.00\\
P527/[8] & \color{red}{0.68} & \color{red}{P31/[10]/0.04} & \color{red}{P641/[10]/0.03} & \color{red}{0.64} & \color{red}{P31/[10]/0.14} & \color{red}{P159/[10]/0.06} & {0.71} & {P31/[10]/0.07} & {P159/[10]/0.06}\\
P155/[8] & 0.86 & P176/[10]/0.01 & P31/[10]/0.01 & 0.79 & P31/[10]/0.09 & P176/[10]/0.04 & 0.86 & P176/[10]/0.02 & P31/[10]/0.01\\
P27/[8] & 0.91 & P407/[10]/0.02 & P937/[7]/0.01 & 0.81 & P407/[10]/0.15 & P17/[6]/0.01 & 0.94 & P407/[10]/0.02 & P137/[7]/0.01\\
P58/[8] & \color{red}{0.63} & \color{red}{P1877/[10]/0.18} & \color{red}{P57/[8]/0.14} & \color{red}{0.24} & \color{red}{P1877/[10]/0.69} & \color{red}{P57/[8]/0.06} & \color{red}{0.35} & \color{red}{P1877/[10]/0.51} & \color{red}{P57/[8]/0.10}\\
P57/[8] & 0.87 & P1877/[10]/0.04 & P58/[8]/0.04 & \color{red}{0.59} & \color{red}{P1877/[10]/0.36} & \color{red}{P58/[8]/0.03} & 0.82 & P1877/[10]/0.14 & P58/[8]/0.01\\
\midrule
P403/[9] & 0.81 & P974/[7]/0.16 & P4552/[6]/0.01 & 0.95 & P974/[7]/0.04 & P17/[6]/0.01 & 0.98 & P974/[7]/0.01 & P159/[10]/0.01\\
P306/[9] & 0.96 & P400/[3]/0.02 & P176/[10]/0.01 & 0.96 & P31/[10]/0.02 & P176/[10]/0.01 & 0.99 & P176/[10]/0.01 & P400/[3]/0.01\\
P175/[9] & 0.93 & P86/[5]/0.05 & P57/[8]/0.01 & 0.96 & P1877/[10]/0.02 & P176/[10]/0.01 & 0.96 & P1877/[10]/0.01 & P57/[8]/0.01\\
P102/[9] & 0.97 & P463/[5]/0.01 & P551/[7]/0.01 & 0.97 & P937/[7]/0.01 & P1344/[10]/0.01 & 0.96 & P463/[5]/0.01 & P937/[7]/0.01\\
P1411/[9] & 1.00 & P137/[7]/0.00 & P3450/[7]/0.00 & 0.99 & P31/[10]/0.01 & P921/[7]/0.00 & 1.00 & P137/[7]/0.00 & P3450/[7]/0.00\\
P40/[9] & 0.86 & P22/[10]/0.05 & P26/[8]/0.04 & 0.84 & P22/[10]/0.15 & P26/[8]/0.01 & 0.93 & P22/[10]/0.06 & P3373/[7]/0.01\\
P105/[9] & 0.99 & P31/[10]/0.01 & P137/[7]/0.00 & 0.98 & P31/[10]/0.02 & P137/[7]/0.00 & 0.99 & P31/[10]/0.01 & P137/[7]/0.00\\
P460/[9] & {0.71} & {P31/[10]/0.06} & {P155/[8]/0.03} & \color{red}{0.62} & \color{red}{P31/[10]/0.24} & \color{red}{P22/[10]/0.02} & 0.76 & P31/[10]/0.09 & P135/[6]/0.01\\
\midrule
P176/[10] & 0.95 & P178/[7]/0.03 & P127/[5]/0.01 & 0.96 & P159/[10]/0.01 & P355/[2]/0.01 & 0.96 & P178/[7]/0.01 & P159/[10]/0.01\\
P641/[10] & 0.98 & P106/[4]/0.01 & P101/[3]/0.01 & 0.99 & P106/[4]/0.01 & P937/[7]/0.00 & 0.99 & P106/[4]/0.01 & P937/[7]/0.00\\
P22/[10] & 0.89 & P3373/[7]/0.04 & P40/[9]/0.02 & 0.98 & P175/[9]/0.01 & P3373/[7]/0.01 & 0.95 & P40/[9]/0.02 & P175/[9]/0.01\\
P31/[10] & 0.81 & P1411/[9]/0.04 & P306/[9]/0.03 & 0.99 & P361/[2]/0.01 & P178/[7]/0.00 & 0.94 & P136/[6]/0.01 & P463/[5]/0.01\\
P1344/[10] & 0.97 & P2094/[3]/0.01 & P118/[2]/0.01 & 0.99 & P2094/[3]/0.01 & P3450/[7]/0.00 & 0.98 & P2094/[3]/0.01 & P40/[9]/0.01\\
P407/[10] & 0.92 & P364/[3]/0.04 & P27/[8]/0.02 & 0.99 & P31/[10]/0.01 & P137/[7]/0.00 & 0.97 & P136/[6]/0.01 & P27/[8]/0.01\\
P1877/[10] & 0.86 & P58/[8]/0.13 & P57/[8]/0.01 & 0.99 & P58/[8]/0.01 & P937/[7]/0.00 & 0.96 & P58/[8]/0.02 & P57/[8]/0.01\\
P159/[10] & 0.80 & P740/[5]/0.07 & P551/[7]/0.03 & 0.98 & P150/[3]/0.01 & P1001/[6]/0.01 & 0.94 & P1408/[5]/0.01 & P740/[5]/0.01\\

        \bottomrule
    \end{tabular}
    }
    \caption{Error analysis on \fewrel (relations from task $6$ to task $10$).}
    \label{tab:error2}
\end{table*}

\end{document}